\documentclass{article}

\PassOptionsToPackage{numbers}{natbib}

\usepackage[preprint]{neurips_2020}

\usepackage[utf8]{inputenc} 
\usepackage[T1]{fontenc}    
\usepackage{hyperref}       
\usepackage{url}            
\usepackage{booktabs}       
\usepackage{amsfonts}       
\usepackage{nicefrac}       
\usepackage{microtype}      
\usepackage{array}
\usepackage{pbox}
\usepackage{bm}
\usepackage{amsmath}
\usepackage{amssymb}
\usepackage{subcaption}
\usepackage{booktabs} 
\usepackage{multirow}
\usepackage{floatrow}
\usepackage{float}
\floatstyle{plaintop}
\restylefloat{table}
\newfloatcommand{capbtabbox}{table}[][\FBwidth]
\usepackage{listings}
\usepackage{graphicx}
\DeclareMathOperator{\diag}{diag}

\usepackage{wrapfig}

\title{Improving Sequential Latent Variable Models\\ with Autoregressive Flows}

\author{
Joseph Marino\\
California Institute of Technology\\
\texttt{jmarino@caltech.edu}
\And
Lei Chen\\
Simon Fraser University\\
\texttt{lei\_chen\_4@sfu.ca}
\AND
Jiawei He\\
Simon Fraser University\\
\texttt{jha203@sfu.ca}
\And
Stephan Mandt\\
University of California Irvine\\
\texttt{mandt@uci.edu}
}

\begin{document}

\maketitle

\begin{abstract}
We propose an approach for improving sequence modeling based on autoregressive normalizing flows. Each autoregressive transform, acting across time, serves as a moving frame of reference, removing temporal correlations and simplifying the modeling of higher-level dynamics. This technique provides a simple, general-purpose method for improving sequence modeling, with connections to existing and classical techniques. We demonstrate the proposed approach both with standalone flow-based models and as a component within sequential latent variable models. Results are presented on three benchmark video datasets and three other time series datasets, where autoregressive flow-based dynamics improve log-likelihood performance over baseline models. Finally, we illustrate the decorrelation and improved generalization properties of using flow-based dynamics.
\end{abstract}

\section{Introduction}
\label{sec: introduction}

Data often contain sequential structure, providing a rich signal for learning models of the world. Such models are useful for representing sequences \citep{li2018deep,ha2018recurrent} and planning actions \citep{hafner2019learning,chua2018deep}. Recent advances in deep learning have facilitated learning sequential probabilistic models directly from high-dimensional data \citep{graves2013generating}, like audio and video. A variety of techniques have emerged for learning deep sequential models, including memory units \citep{hochreiter1997long} and stochastic latent variables \citep{chung2015recurrent,bayer2014learning}. These techniques have enabled sequential models to capture increasingly complex dynamics. In this paper, we explore the complementary direction, asking \textit{can we simplify the dynamics of the data to meet the capacity of the model?} To do so, we aim to learn a \textit{frame of reference} to assist in modeling the data.

Frames of reference are an important consideration in sequence modeling, as they can simplify dynamics by removing redundancy. For instance, in a physical system, the frame of reference that moves with the system's center of mass removes the redundancy in displacement. Frames of reference are also more widely applicable to arbitrary sequences. Indeed, video compression schemes use predictions as a frame of reference to remove temporal redundancy \citep{oliver1952efficient,agustsson2020scale, yang2021hierarchical}. By learning and applying a similar type of temporal normalization for sequence modeling, the model can focus on aspects that are not predicted by the low-level frame of reference, thereby simplifying dynamics modeling.

\begin{figure}[t]
    \centering
    \includegraphics[width=\textwidth]{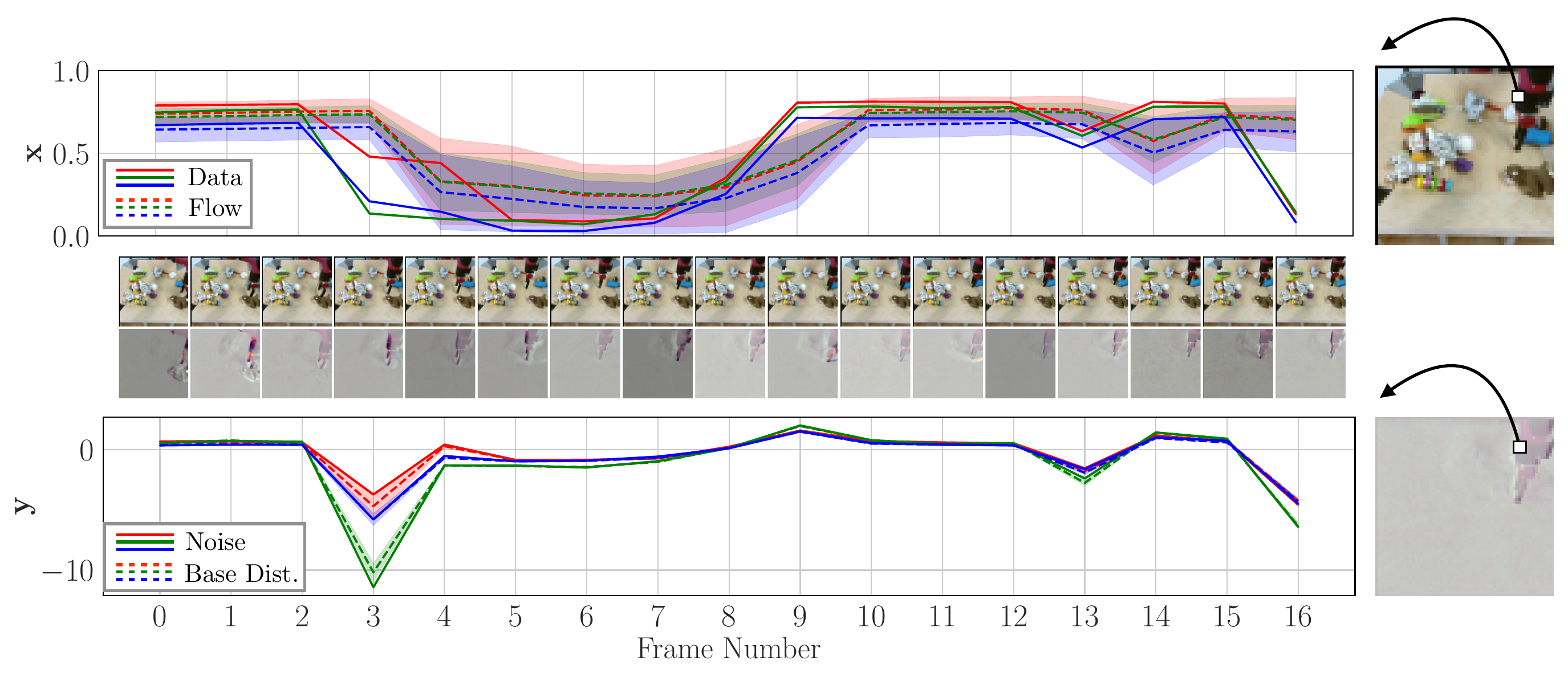}
    \caption{\textbf{Sequence Modeling with Autoregressive Flows}. \textbf{Top}: Pixel values (solid) for a particular pixel location in a video sequence. An autoregressive flow models the pixel sequence using an affine shift (dashed) and scale (shaded), acting as a frame of reference. \textbf{Middle}: Frames of the data sequence (top) and the resulting ``noise'' (bottom) from applying the shift and scale. The redundant, static background has been largely removed. \textbf{Bottom}: The noise values (solid) are modeled using a base distribution (dashed and shaded) provided by a higher-level model. By removing temporal redundancy from the data sequence, the autoregressive flow simplifies dynamics modeling.}
    \label{fig:data vs noise}
\end{figure}

We formalize this notion of temporal normalization through the framework of autoregressive normalizing flows \citep{kingma2016improved,papamakarios2017masked}. In the context of sequences, these flows form predictions across time, attempting to remove temporal dependencies \citep{srinivasan1982predictive}. Thus, autoregressive flows can act as a pre-processing technique to simplify dynamics. We preview this approach in Figure~\ref{fig:data vs noise}, where an autoregressive flow modeling the data (top) creates a transformed space for modeling dynamics (bottom). The transformed space is largely invariant to absolute pixel value, focusing instead on capturing deviations and motion.

We empirically demonstrate this modeling technique, both with standalone autoregressive normalizing flows, as well as within sequential latent variable models. While normalizing flows have been applied in sequential contexts previously, our main contributions are in \textbf{1)} showing how these models can act as a general pre-processing technique to improve dynamics modeling, \textbf{2)} empirically demonstrating log-likelihood performance improvements, as well as generalization improvements, on three benchmark video datasets and time series data from the UCI machine learning repository. This technique also connects to previous work in dynamics modeling, probabilistic models, and sequence compression, enabling directions for further investigation.

\section{Background}
\label{sec: background}

\subsection{Autoregressive Models}
\label{sec: ar models}

Consider modeling discrete sequences of observations, $\mathbf{x}_{1:T} \sim p_{\small \textrm{data}} (\mathbf{x}_{1:T})$, using a probabilistic model, $p_\theta (\mathbf{x}_{1:T})$, with parameters $\theta$. Autoregressive models \citep{frey1996does,bengio2000modeling} use the chain rule of probability to express the joint distribution over time steps as the product of $T$ conditional distributions. These models are often formulated in forward temporal order:
\begin{equation}
    p_\theta (\mathbf{x}_{1:T}) = \prod_{t=1}^T p_\theta (\mathbf{x}_t | \mathbf{x}_{<t}).
    \label{eq: autoregression}
\end{equation}
Each conditional distribution, $p_\theta (\mathbf{x}_t | \mathbf{x}_{<t})$, models the dependence between time steps. For continuous variables, it is often assumed that each distribution takes a simple form, such as a diagonal Gaussian: $p_\theta (\mathbf{x}_t | \mathbf{x}_{<t}) = \mathcal{N} (\mathbf{x}_t ; \bm{\mu}_\theta (\mathbf{x}_{<t}), \diag (\bm{\sigma}^2_\theta (\mathbf{x}_{<t}))),$
where $\bm{\mu}_\theta (\cdot)$ and $\bm{\sigma}_\theta (\cdot)$ are functions denoting the mean and standard deviation. These functions may take past observations as input through a recurrent network or a convolutional window \citep{van2016wavenet}. When applied to spatial data \citep{van2016pixel}, autoregressive models excel at capturing local dependencies. However, due to their restrictive forms, such models often struggle to capture more complex structure.

\subsection{Autoregressive (Sequential) Latent Variable Models}
\label{sec: ar lvms}

Autoregressive models can be improved by incorporating latent variables \citep{murphy2012machine}, often represented as a corresponding sequence, $\mathbf{z}_{1:T}$. The joint distribution, $p_\theta (\mathbf{x}_{1:T} , \mathbf{z}_{1:T})$, has the form:
\begin{equation}
    p_\theta (\mathbf{x}_{1:T} , \mathbf{z}_{1:T}) = \prod_{t=1}^T p_\theta (\mathbf{x}_t | \mathbf{x}_{<t}, \mathbf{z}_{\leq t}) p_\theta (\mathbf{z}_t | \mathbf{x}_{<t} , \mathbf{z}_{<t}).
\end{equation}
Unlike the Gaussian form, evaluating $p_\theta (\mathbf{x}_t | \mathbf{x}_{<t})$ now requires integrating over the latent variables,
\begin{equation}
    p_\theta (\mathbf{x}_t | \mathbf{x}_{<t}) = \int p_\theta (\mathbf{x}_t | \mathbf{x}_{<t}, \mathbf{z}_{\leq t}) p_\theta (\mathbf{z}_{\leq t} | \mathbf{x}_{<t}) d \mathbf{z}_{\leq t},
\end{equation} 
yielding a more flexible distribution. However, performing this integration in practice is typically intractable, requiring approximate inference techniques, like variational inference \citep{jordan1998introduction}, or invertible models \citep{kumar2019videoflow}. Recent works have parameterized these models with deep neural networks, e.g. \citep{chung2015recurrent,gan2015deep,fraccaro2016sequential,karl2017deep}, using amortized variational inference \citep{kingma2014stochastic,rezende2014stochastic}. Typically, the conditional likelihood, $p_\theta (\mathbf{x}_t | \mathbf{x}_{<t}, \mathbf{z}_{\leq t})$, and the prior, $p_\theta (\mathbf{z}_t | \mathbf{x}_{<t} , \mathbf{z}_{<t})$, are Gaussian densities, with temporal conditioning handled through recurrent networks. Such models have demonstrated success in audio \citep{chung2015recurrent,fraccaro2016sequential} and video modeling \citep{xue2016visual,gemici2017generative,denton2018stochastic,he2018probabilistic,li2018deep}. However, as noted by \cite{kumar2019videoflow}, such models can be difficult to train with standard log-likelihood objectives, often struggling to capture dynamics.

\begin{figure}[t!]
    \centering
    \includegraphics[width=0.55\textwidth]{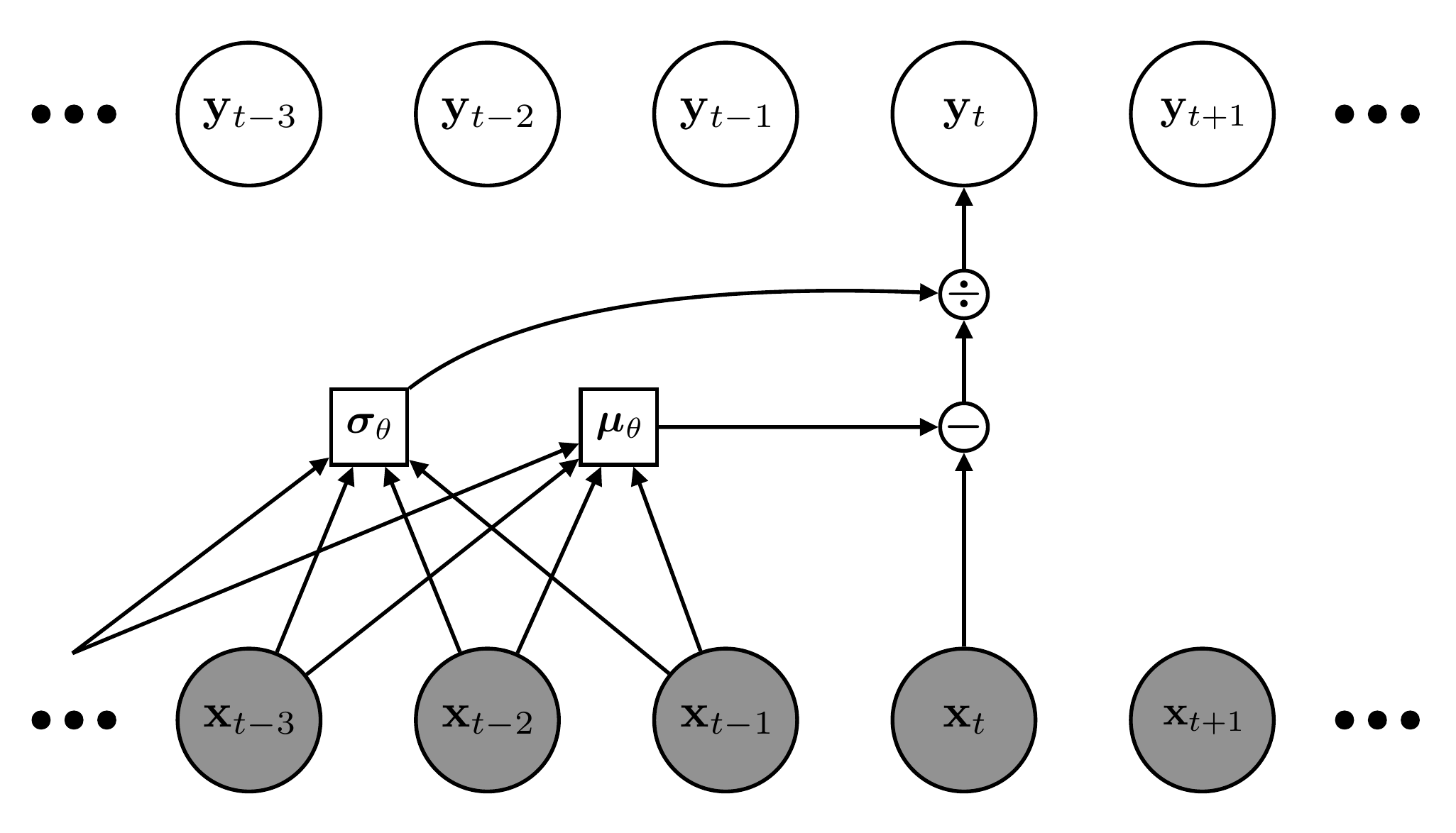}
    \caption{\textbf{Affine Autoregressive Transform}. Computational diagram for an affine autoregressive transform \cite{papamakarios2017masked}. Each $\mathbf{y}_t$ is an affine transform of $\mathbf{x}_t$, with the affine parameters potentially non-linear functions of $\mathbf{x}_{<t}$. The inverse transform, shown here, is capable of converting a \textit{correlated} input, $\mathbf{x}_{1:T}$, into an \textit{uncorrelated} output, $\mathbf{y}_{1:T}$.}
    \label{fig: autoregressive transforms}
\end{figure}

\subsection{Autoregressive Flows}

Our approach is based on affine autoregressive normalizing flows \citep{kingma2016improved,papamakarios2017masked}. Here, we continue with the perspective of temporal sequences, however, these flows were initially developed and demonstrated in \textit{static} settings. \cite{kingma2016improved} noted that sampling from an autoregressive Gaussian model is an invertible transform, resulting in a \textit{normalizing flow} \citep{rippel2013high,dinh2015nice,dinh2017density,rezende2015variational}. Flow-based models transform simple, \textit{base} probability distributions into more complex ones while maintaining exact likelihood evaluation. To see their connection to autoregressive models, we can express sampling a Gaussian random variable using the reparameterization trick \citep{kingma2014stochastic,rezende2014stochastic}:
\begin{equation}
    \mathbf{x}_t = \bm{\mu}_\theta (\mathbf{x}_{<t}) + \bm{\sigma}_\theta (\mathbf{x}_{<t}) \odot \mathbf{y}_t,
    \label{eq: autoreg reparam}
\end{equation}
where $\mathbf{y}_t \sim \mathcal{N} (\mathbf{y}_t; \mathbf{0}, \mathbf{I})$ is an auxiliary random variable and $\odot$ denotes element-wise multiplication. Thus, $\mathbf{x}_t$ is an invertible transform of $\mathbf{y}_t$, with the inverse given as
\begin{equation}
    \mathbf{y}_t = \frac{\mathbf{x}_t - \bm{\mu}_\theta (\mathbf{x}_{<t})}{\bm{\sigma}_\theta (\mathbf{x}_{<t})},
    \label{eq: autoreg reparam inv}
\end{equation}
where division is element-wise. The inverse transform in Eq.\ \ref{eq: autoreg reparam inv}, shown in Figure~\ref{fig: autoregressive transforms}, normalizes (hence, \textit{normalizing} flow) $\mathbf{x}_{1:T}$, removing statistical dependencies. Given the functional mapping between $\mathbf{y}_t$ and $\mathbf{x}_t$ in Eq.\ \ref{eq: autoreg reparam}, the change of variables formula converts between probabilities in each space:
\begin{equation}
    \log p_\theta (\mathbf{x}_{1:T}) = \log p_\theta (\mathbf{y}_{1:T}) - \log \left| \det \left( \frac{\partial \mathbf{x}_{1:T}}{\partial \mathbf{y}_{1:T}} \right) \right|.
    \label{eq: change of variables 1}
\end{equation}
By the construction of Eqs. \ref{eq: autoreg reparam} and \ref{eq: autoreg reparam inv}, the Jacobian in Eq. \ref{eq: change of variables 1} is triangular, enabling efficient evaluation as the product of diagonal terms:
\begin{equation}
    \log \left| \det \left( \frac{\partial \mathbf{x}_{1:T}}{\partial \mathbf{y}_{1:T}} \right) \right| = \sum_{t=1}^T \sum_i \log \sigma_{\theta, i} (\mathbf{x}_{<t}),
\end{equation}
where $i$ denotes the observation dimension, e.g. pixel. For a Gaussian autoregressive model, the base distribution is $p_\theta (\mathbf{y}_{1:T}) = \mathcal{N} (\mathbf{y}_{1:T}; \mathbf{0}, \mathbf{I})$. We can improve upon this simple set-up by chaining transforms together, i.e. parameterizing $p_\theta (\mathbf{y}_{1:T})$ as a flow, resulting in hierarchical models.

\subsection{Related Work}
\label{sec: related work}

Autoregressive flows were initially considered in the contexts of variational inference \citep{kingma2016improved} and generative modeling \citep{papamakarios2017masked}. These approaches are generalizations of previous approaches with affine transforms \citep{dinh2015nice,dinh2017density}. While autoregressive flows are well-suited for sequential data, these approaches, as well as many recent approaches \citep{huang2018neural,oliva2018transformation,kingma2018glow}, were initially applied to static data, such as images.

Recent works have started applying flow-based models to sequential data.  \cite{van2018parallel} and \cite{ping2019clarinet} \textit{distill} autoregressive speech models into flow-based models. \cite{prenger2019waveglow} and  \cite{kim2019flowavenet} instead train these models directly.  \cite{kumar2019videoflow} use a flow to model individual video frames, with an autoregressive prior modeling dynamics across time steps.  \cite{rhinehart2018r2p2,rhinehart2019precog} use autoregressive flows for modeling vehicle motion, and  \cite{henter2019moglow} use flows for motion synthesis with motion-capture data.  \cite{ziegler2019latent} model discrete observations (e.g., text) by using flows to model dynamics of continuous latent variables. Like these recent works, we apply flow-based models to sequences. However, we demonstrate that autoregressive flows can serve as a general-purpose technique for improving dynamics models. To the best of our knowledge, our work is the first to use flows to pre-process sequences to improve sequential latent variable models.

We utilize affine flows (Eq.~\ref{eq: autoreg reparam}), a family that includes methods like NICE \citep{dinh2015nice}, RealNVP \citep{dinh2017density}, IAF \citep{kingma2016improved}, MAF \citep{papamakarios2017masked}, and Glow \citep{kingma2018glow}. However, there has been recent work in non-affine flows \citep{huang2018neural,jaini2019sum,durkan2019neural}, which offer further flexibility. We chose to investigate affine flows because they are commonly employed and relatively simple, however, non-affine flows could result in additional improvements.

Autoregressive dynamics models are also prominent in other related areas. Within the statistics and econometrics literature, autoregressive integrated moving average (ARIMA) is a standard technique \citep{box2015time, hamilton2020time}, calculating differences with an autoregressive prediction to remove non-stationary components of a temporal signal. Such methods simplify downstream modeling, e.g., by removing seasonal effects. Low-level autoregressive models are also found in audio \citep{atal1979predictive} and video compression codecs \citep{wiegand2003overview, agustsson2020scale,  yang2021hierarchical}, using predictive coding \citep{oliver1952efficient} to remove temporal redundancy, thereby improving downstream compression rates. Intuitively, if sequential inputs are highly predictable, it is far more efficient to compress the prediction error rather than each input (e.g., video frame) separately. Finally, we note that autoregressive models are a generic dynamics modeling approach and can, in principle, be parameterized by other techniques, such as LSTMs \citep{hochreiter1997long}, or combined with other models, such as hidden Markov models (HMMs) \citep{murphy2012machine}.

\section{Method}
\label{sec: approach}

We now describe our approach for improving sequence modeling. First, we motivate using autoregressive flows to reduce temporal dependencies, thereby simplifying dynamics. We then show how this simple technique can be incorporated within sequential latent variable models.

\begin{figure}[t!]
    \centering
    \begin{subfigure}[t]{0.24\textwidth}
        \centering
        \includegraphics[width=\textwidth]{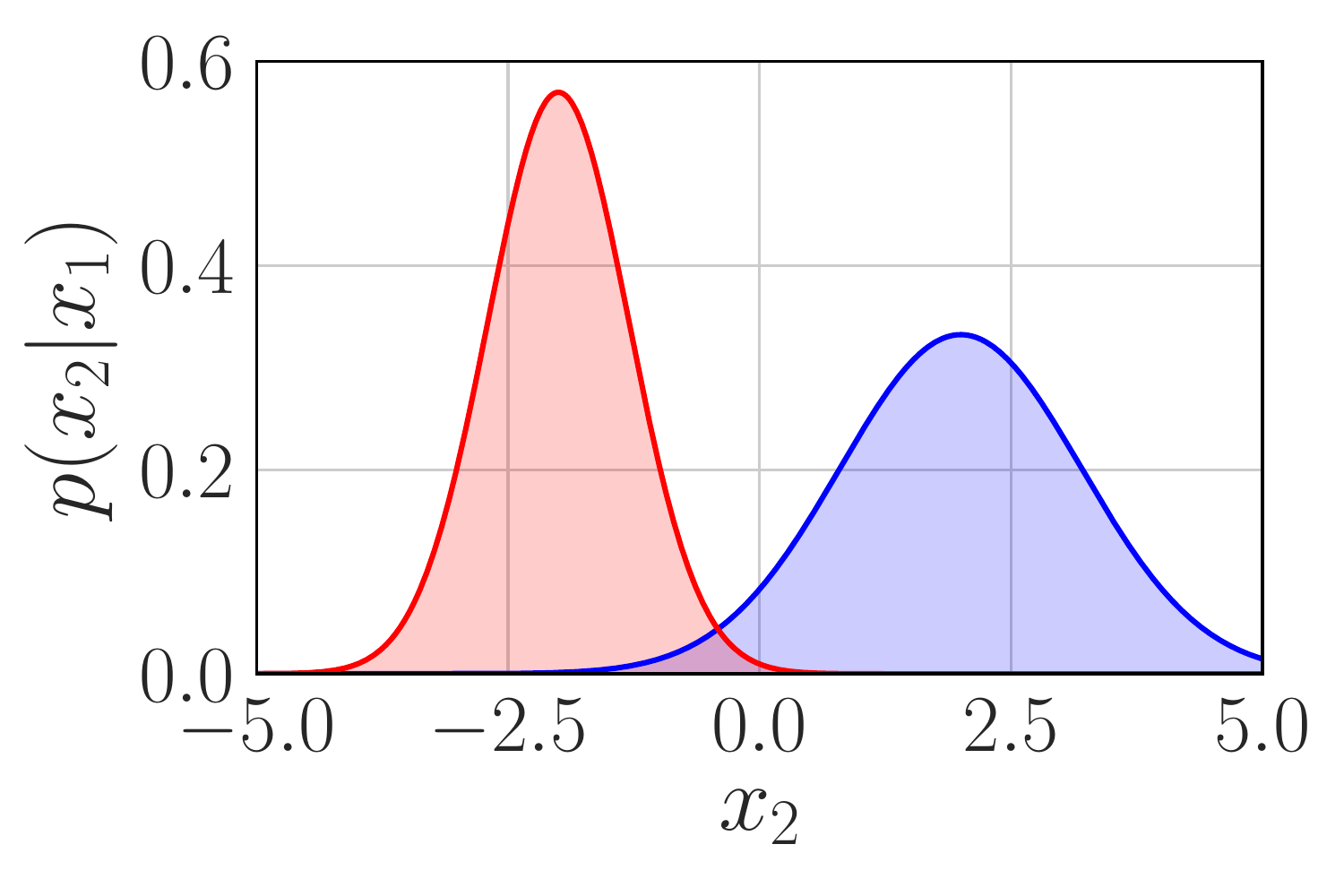}
        \caption{}
        \label{fig: x_cond}
    \end{subfigure}%
    ~
    \begin{subfigure}[t]{0.24\textwidth}
        \centering
        \includegraphics[width=\textwidth]{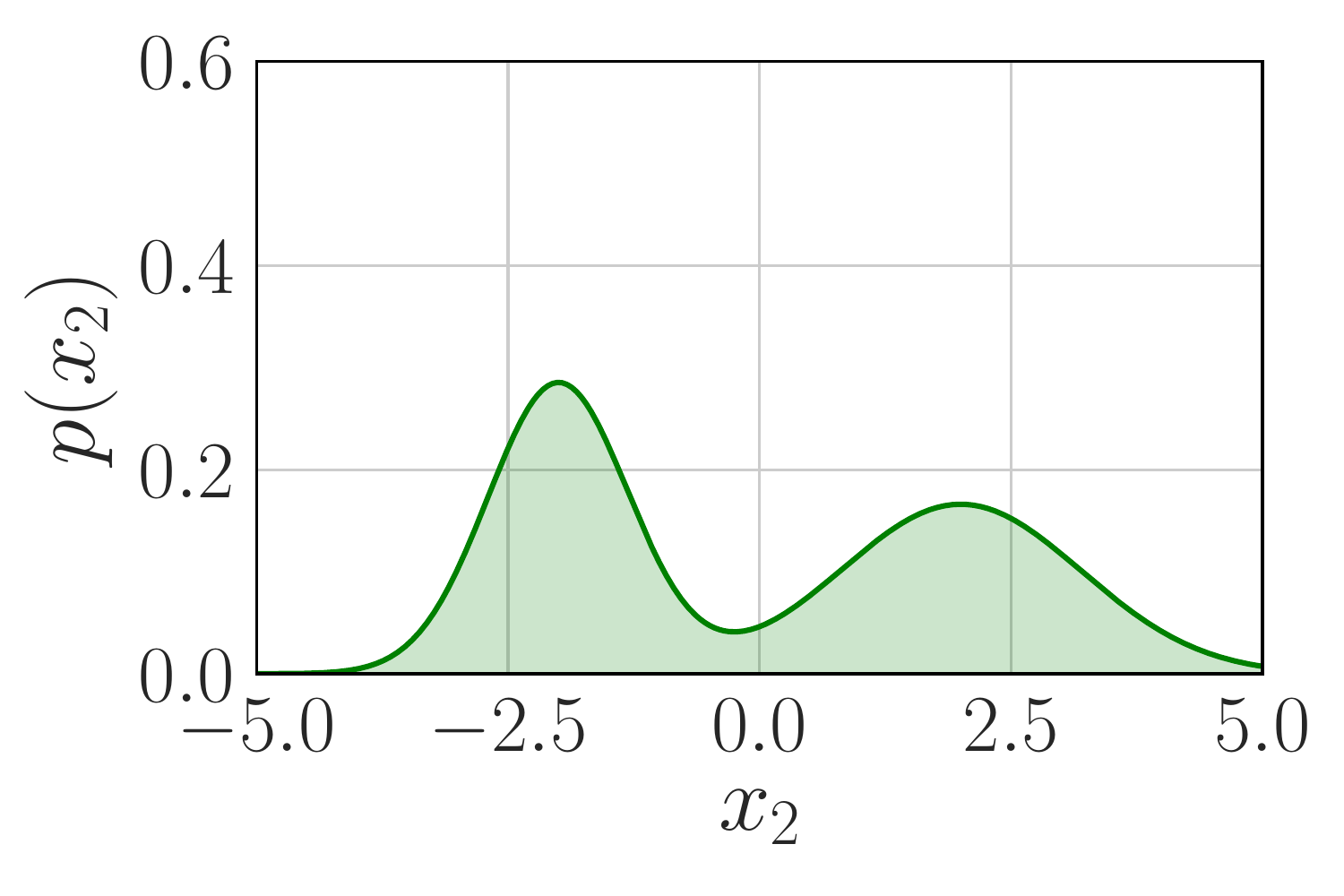}
        \caption{}
        \label{fig: x_marg}
    \end{subfigure}%
    ~
    \begin{subfigure}[t]{0.24\textwidth}
        \centering
        \includegraphics[width=\textwidth]{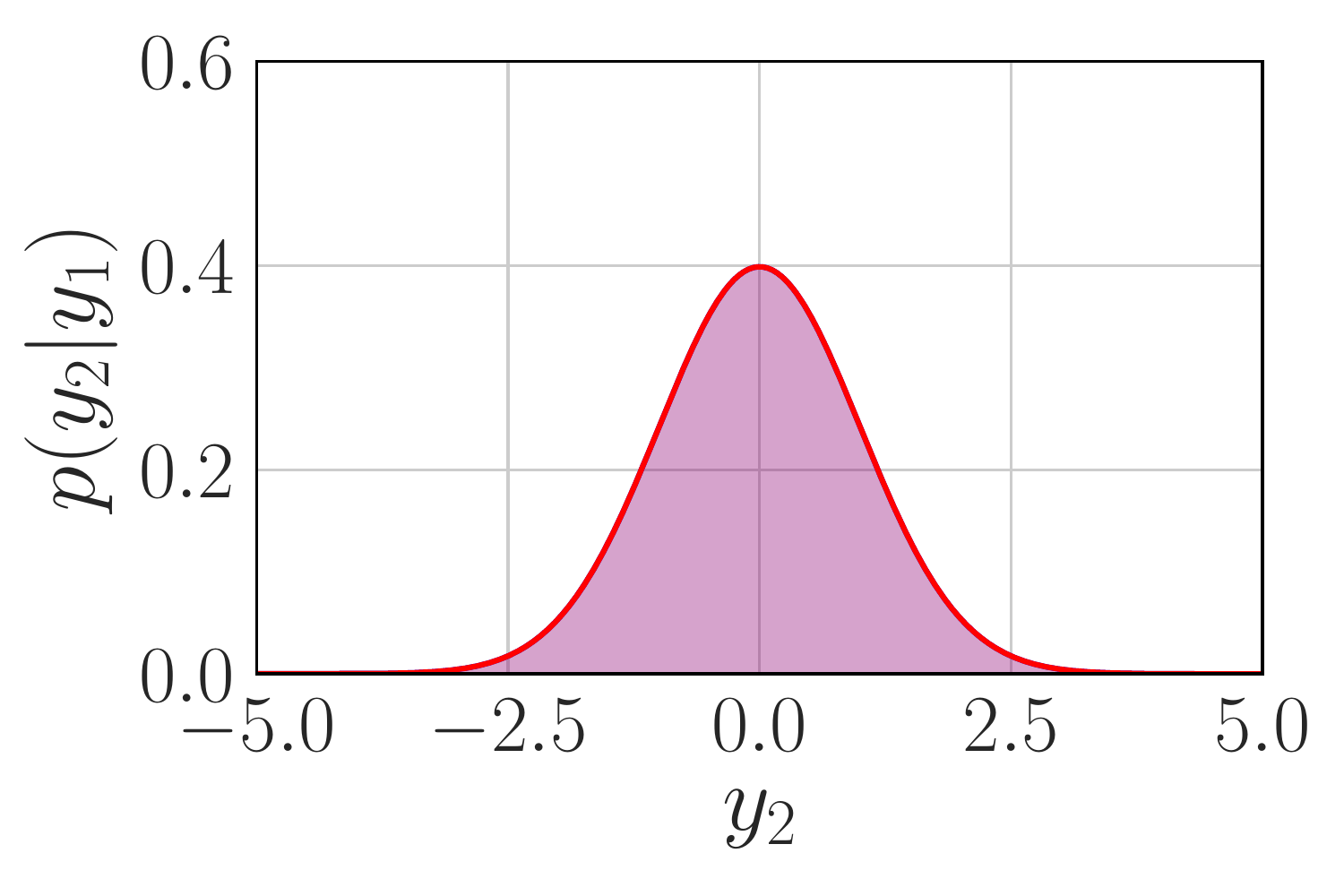}
        \caption{}
        \label{fig: y_cond}
    \end{subfigure}%
    ~
    \begin{subfigure}[t]{0.24\textwidth}
        \centering
        \includegraphics[width=\textwidth]{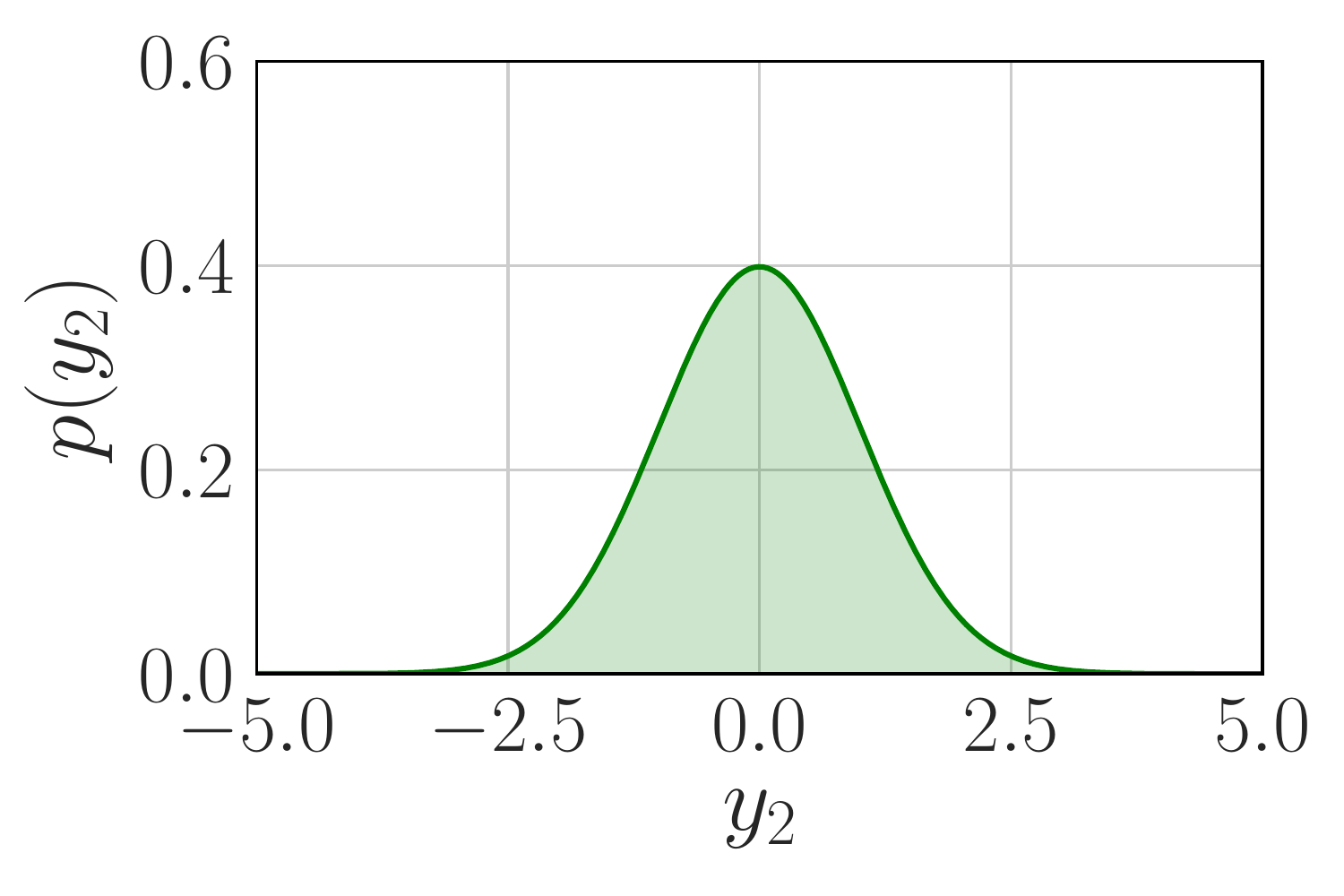}
        \caption{}
        \label{fig: y_marg}
    \end{subfigure}
    \caption{\textbf{Redundancy Reduction}. \textbf{(a)} Conditional densities for $p (x_2 | x_1)$. \textbf{(b)} The marginal, $p (x_2)$ differs from the conditional densities, thus, $\mathcal{I} (x_1 ; x_2) > 0$. \textbf{(c)} In the normalized space of $y$, the corresponding densities $p (y_2 | y_1)$ are identical. \textbf{(d)} The marginal $p(y_2)$ is identical to the conditionals, so $\mathcal{I} (y_1; y_2) = 0.$ Thus, in this case, a conditional affine transform removed the dependencies.}
    \label{fig: redundancy reduction}
\end{figure}

\subsection{Motivation: Temporal Redundancy Reduction}
\label{sec: motivation}

Normalizing flows, while often utilized for density estimation, originated from data pre-processing techniques \citep{friedman1987exploratory,hyvarinen2000independent,chen2001gaussianization}, which remove dependencies between dimensions, i.e., \textit{redundancy reduction} \citep{barlow1961possible}. Removing dependencies simplifies the resulting probability distribution by restricting variation to individual dimensions, generally simplifying downstream tasks \citep{laparra2011iterative}. Normalizing flows improve upon these procedures using flexible, non-linear functions \citep{deco1995higher,dinh2015nice}. While flows have been used for spatial decorrelation \citep{agrawal2016deep,winkler2019learning} and with other models \citep{huang2017learnable}, this capability remains under-explored.

Our main contribution is showing how to utilize autoregressive flows for temporal pre-processing to improve dynamics modeling. Data sequences contain dependencies in time, for example, in the redundancy of video pixels (Figure~\ref{fig:data vs noise}), which are often highly predictable. These dependencies define the \textit{dynamics} of the data, with the degree of dependence quantified by the multi-information, 
\begin{equation}
\mathcal{I} (\mathbf{x}_{1:T}) = \sum_t \mathcal{H} (\mathbf{x}_t) - \mathcal{H} (\mathbf{x}_{1:T}),
\end{equation}
where $\mathcal{H}$ denotes entropy. Normalizing flows are capable of reducing redundancy, arriving at a new sequence, $\mathbf{y}_{1:T}$, with $\mathcal{I} (\mathbf{y}_{1:T}) \leq \mathcal{I} (\mathbf{x}_{1:T})$, thereby reducing temporal dependencies. Thus, rather than fit the data distribution directly, we can first simplify the dynamics by pre-processing sequences with a normalizing flow, then fitting the resulting sequence. Through training, the flow will attempt to remove redundancies to meet the modeling capacity of the higher-level dynamics model, $p_\theta (\mathbf{y}_{1:T})$.

\paragraph{Example} To visualize this procedure for an affine autoregressive flow, consider a one-dimensional input over two time steps, $x_1$ and $x_2$. For each value of $x_1$, there is a conditional density, $p (x_2 | x_1)$. Assume that these densities take one of two forms, which are identical but shifted and scaled, shown in Figure~\ref{fig: redundancy reduction}. Transforming these densities through their conditional means, $\mu_2 = \mathbb{E} \left[ x_2 | x_1 \right]$, and standard deviations, $\sigma_2 = \mathbb{E} \left[ (x_2 - \mu_2)^2 | x_1 \right]^{1/2}$, creates a normalized space, $y_2 = (x_2 - \mu_2) / \sigma_2$, where the conditional densities are identical. In this space, the multi-information is
\begin{equation}
    \nonumber \mathcal{I} (y_1;y_2) = \mathbb{E}_{p (y_1 , y_2)} \left[ \log p (y_2 | y_1) - \log p(y_2) \right] = 0,
\end{equation}
whereas $\mathcal{I} (x_1; x_2) > 0.$ Indeed, if $p (x_t | x_{<t})$ is linear-Gaussian, inverting an affine autoregressive flow exactly corresponds to Cholesky whitening \citep{pourahmadi2011covariance,kingma2016improved}, removing all linear dependencies. 

In the example above, $\mu_2$ and $\sigma_2$ act as a \textit{frame of reference} for estimating $x_2$. More generally, in the special case where $\bm{\mu}_\theta (\mathbf{x}_{<t}) = \mathbf{x}_{t-1}$ and $\bm{\sigma} (\mathbf{x}_{<t}) = \mathbf{1}$, we recover $\mathbf{y}_t = \mathbf{x}_t - \mathbf{x}_{t-1} = \Delta \mathbf{x}_t$. Modeling finite differences (or \textit{generalized coordinates} \citep{friston2008hierarchical})  is a well-established technique, (see, e.g. \citep{chua2018deep,kumar2019videoflow}), which is generalized by affine autoregressive flows.

\begin{figure}[t!]
    \centering
    \begin{subfigure}[t]{0.54\textwidth}
        \centering
        \includegraphics[width=\textwidth]{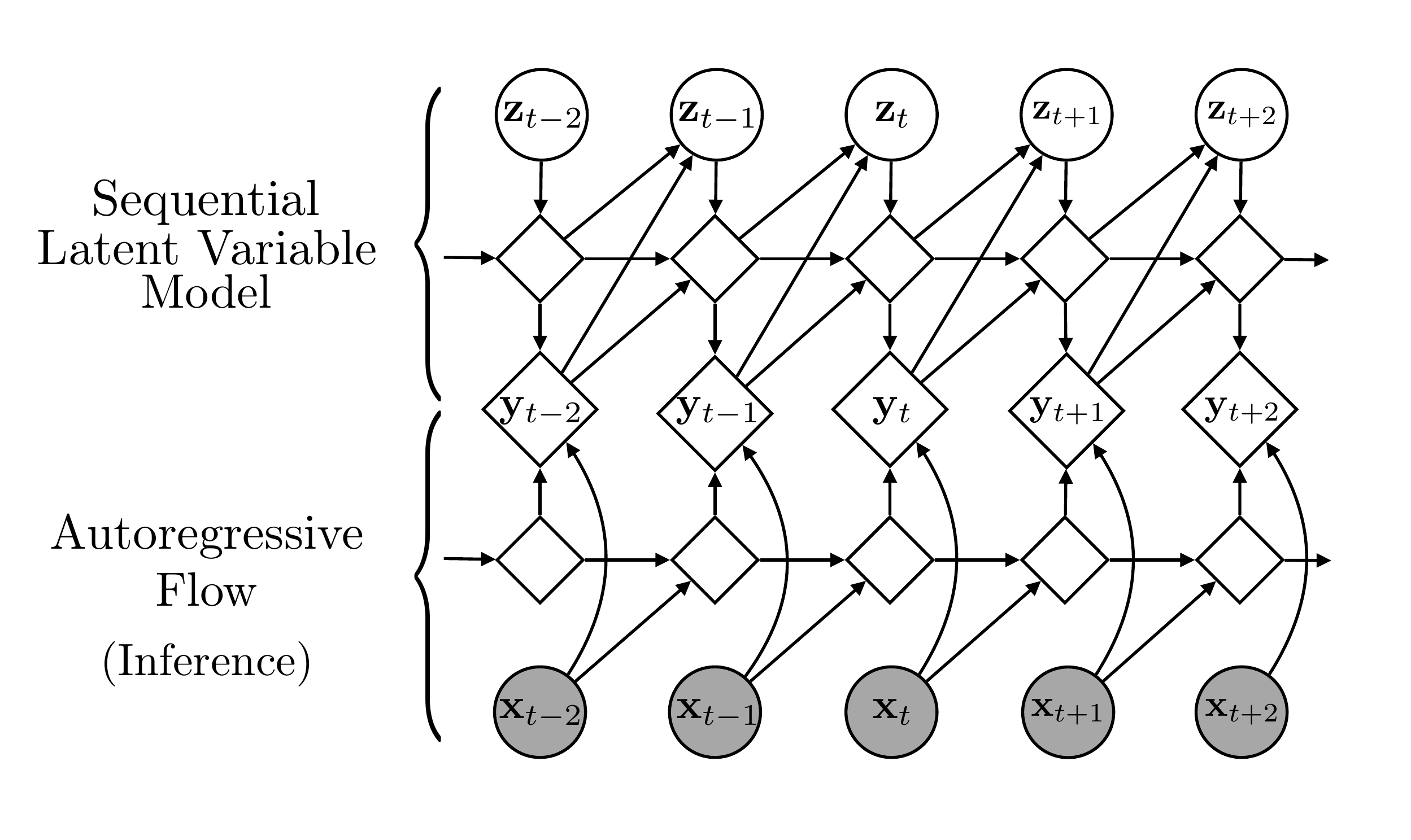}
        \caption{}
        \label{fig: model diagram}
    \end{subfigure}%
    ~ 
    \begin{subfigure}[t]{0.4\textwidth}
        \centering
        \includegraphics[width=\textwidth]{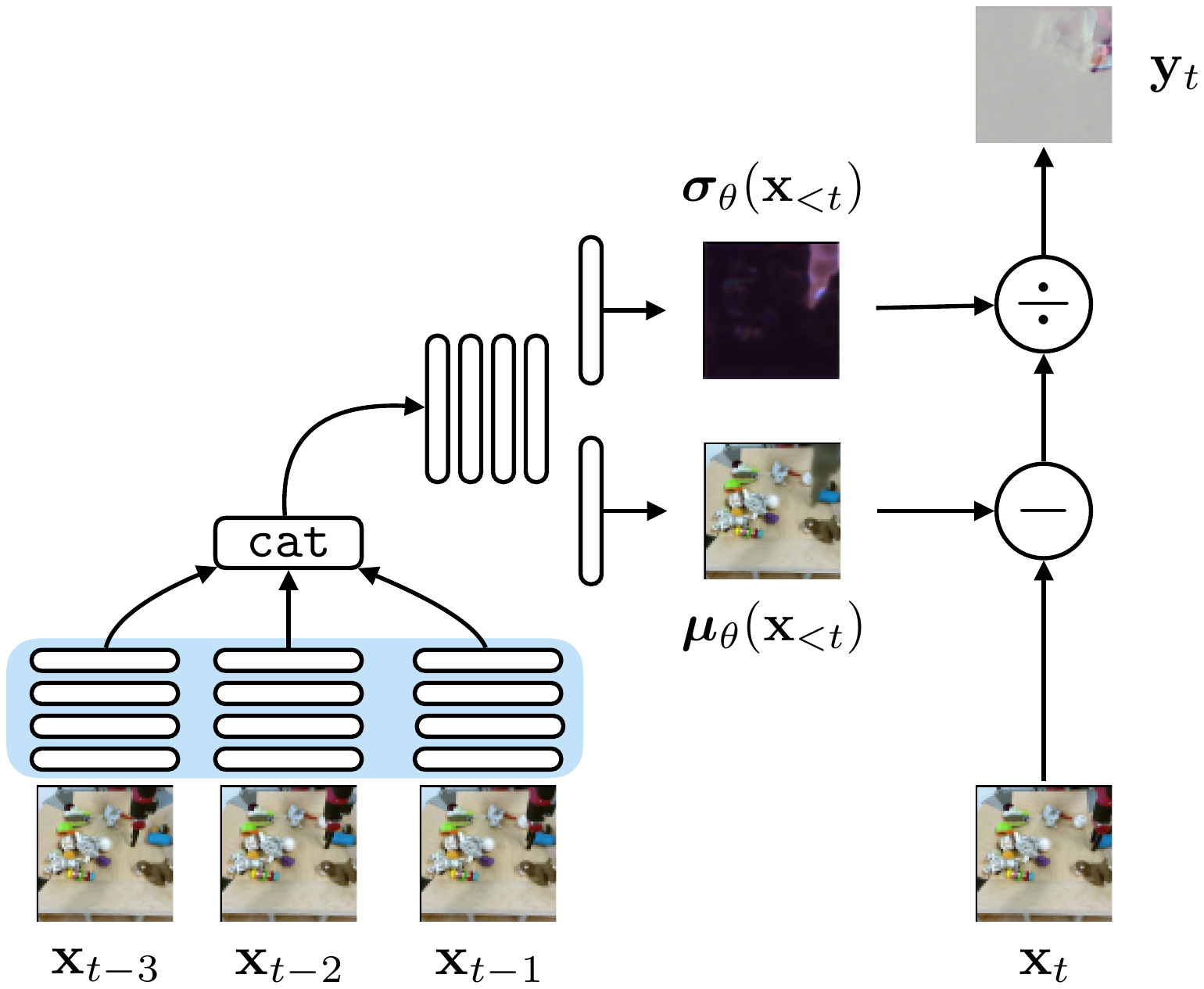}
        \caption{}
        \label{fig: flow architecture}
    \end{subfigure}%
    \caption{\textbf{Model Diagrams}. \textbf{(a)} An autoregressive flow pre-processes a data sequence, $\mathbf{x}_{1:T}$, to produce a new sequence, $\mathbf{y}_{1:T}$, with reduced temporal dependencies. This simplifies dynamics modeling for a higher-level sequential latent variable model, $p_\theta (\mathbf{y}_{1:T}, \mathbf{z}_{1:T})$. Empty diamond nodes represent deterministic dependencies, not recurrent states. \textbf{(b)} Diagram of the autoregressive flow architecture. Blank white rectangles represent convolutional layers (see Appendix). The three stacks of convolutional layers within the blue region are shared. \texttt{cat} denotes channel-wise concatenation.}
    \label{fig: model diagrams}
\end{figure}

\subsection{Modeling Dynamics with Autoregressive Flows}
\label{sec: lvms with af}

We now discuss utilizing autoregressive flows to improve sequence modeling, highlighting use cases for modeling dynamics in the data and latent spaces.

\subsubsection{Data Dynamics}

The form of an affine autoregressive flow across sequences is given in Eqs.~\ref{eq: autoreg reparam} and \ref{eq: autoreg reparam inv}, again, equivalent to a Gaussian autoregressive model. We can stack hierarchical chains of flows to improve the model capacity. Denoting the shift and scale functions at the $m^\textrm{th}$ transform as $\bm{\mu}_\theta^m (\cdot)$ and $\bm{\sigma}_\theta^m (\cdot)$ respectively, we then calculate $\mathbf{y}^m$ using the inverse transform:
\begin{equation}
    \mathbf{y}^m_t = \frac{\mathbf{y}^{m-1}_t - \bm{\mu}^m_\theta (\mathbf{y}^{m-1}_{<t})}{\bm{\sigma}^m_\theta (\mathbf{y}^{m-1}_{<t})}.
\label{eq:inversetransform}
\end{equation}
After the final ($M^\textrm{th}$) transform, we can choose the form of the base distribution, $p_\theta (\mathbf{y}^M_{1:T})$, e.g. Gaussian. While we could attempt to model $\mathbf{x}_{1:T}$ completely using stacked autoregressive flows, these models are limited to affine element-wise transforms that maintain the data dimensionality. Due to this limited capacity, purely flow-based models often require many transforms to be effective \citep{kingma2018glow}.

Instead, we can model the base distribution using an expressive sequential latent variable model (SLVM), or, equivalently, we can augment the conditional likelihood of a SLVM using autoregressive flows (Fig.~\ref{fig: model diagram}). Following the motivation from Section~\ref{sec: motivation}, the flow can remove temporal dependencies, simplifying the modeling task for the SLVM. With a single flow, the joint probability is
\begin{equation}
    p_\theta (\mathbf{x}_{1:T}, \mathbf{z}_{1:T}) = p_\theta (\mathbf{y}_{1:T}, \mathbf{z}_{1:T}) \left| \det \left( \frac{\partial \mathbf{x}_{1:T}}{\partial \mathbf{y}_{1:T}} \right) \right|^{-1},
    \label{eq: change of variables lvm}
\end{equation}
where the SLVM distribution is given by

\begin{equation}
p_\theta (\mathbf{y}_{1:T} , \mathbf{z}_{1:T}) = \prod_{t=1}^T p_\theta (\mathbf{y}_t | \mathbf{y}_{<t}, \mathbf{z}_{\leq t}) p_\theta (\mathbf{z}_t | \mathbf{y}_{<t} , \mathbf{z}_{<t}).
\end{equation}
If the SLVM is itself a flow-based model, we can use maximum log-likelihood training. If not, we can resort to variational inference \citep{chung2015recurrent,fraccaro2016sequential,marino2018general}. We derive and discuss this procedure in the Appendix.

\subsubsection{Latent Dynamics}

We can also consider simplifying latent dynamics modeling using autoregressive flows. This is relevant in hierarchical SLVMs, such as VideoFlow \citep{kumar2019videoflow}, where each latent variable is modeled as a function of past and higher-level latent variables. Using $\mathbf{z}_t^{(\ell)}$ to denote the latent variable at the $\ell^{\textrm{th}}$ level at time $t$, we can parameterize the prior as
\begin{equation}
    p_\theta (\mathbf{z}^{(\ell)}_t | \mathbf{z}^{(\ell)}_{<t}, \mathbf{z}^{(>\ell)}_t) = p_\theta (\mathbf{u}^{(\ell)}_t | \mathbf{u}^{(\ell)}_{<t}, \mathbf{z}^{(>\ell)}_t) \left| \det \left( \frac{\partial \mathbf{z}^{(\ell)}_t}{\partial \mathbf{u}^{(\ell)}_t} \right) \right|^{-1},
\end{equation}
converting $\mathbf{z}^{(\ell)}_t$ into $\mathbf{u}^{(\ell)}_t$ using the inverse transform $\mathbf{u}^{(\ell)}_t = (\mathbf{z}^{(\ell)}_t - \bm{\alpha}_\theta (\mathbf{z}^{(\ell)}_{<t})) / \bm{\beta}_\theta (\mathbf{z}^{(\ell)}_{<t})$. As noted previously, VideoFlow uses a special case of this procedure, setting $\bm{\alpha}_\theta (\mathbf{z}^{(\ell)}_{<t}) = \mathbf{z}^{(\ell)}_{t-1}$ and $\bm{\beta}_\theta (\mathbf{z}^{(\ell)}_{<t}) = \mathbf{1}$. Generalizing this procedure further simplifies dynamics throughout the model.

\section{Evaluation}
\label{sec: Evaluation}

We demonstrate and evaluate the proposed technique on three benchmark video datasets: Moving MNIST \citep{srivastava2015unsupervised}, KTH Actions \citep{schuldt2004recognizing}, and BAIR Robot Pushing \citep{ebert2017self}.
In addition, we also perform experiments on several non-video sequence datasets from the UC Irvine Machine Learning Repository.\footnote{\texttt{https://archive.ics.uci.edu/ml/index.php}} Specifically, we look at an activity recognition dataset (\texttt{activity\_rec}) \citep{palumbo2016human}, an indoor localization dataset (\texttt{smartphone\_sensor}) \citep{barsocchi2016multisource}, and a facial expression recognition dataset (\texttt{facial\_exp}) \citep{de2014grammatical}.
Experimental setups are described in Section~\ref{sec:exp_setup}, followed by a set of analyses in Section~\ref{sec:exp_qualitative}. Further details and results can be found in the Appendix.

\begin{figure}[t!]
    \centering
    \begin{subfigure}[t]{0.68\textwidth}
    \centering
    \includegraphics[width=\textwidth]{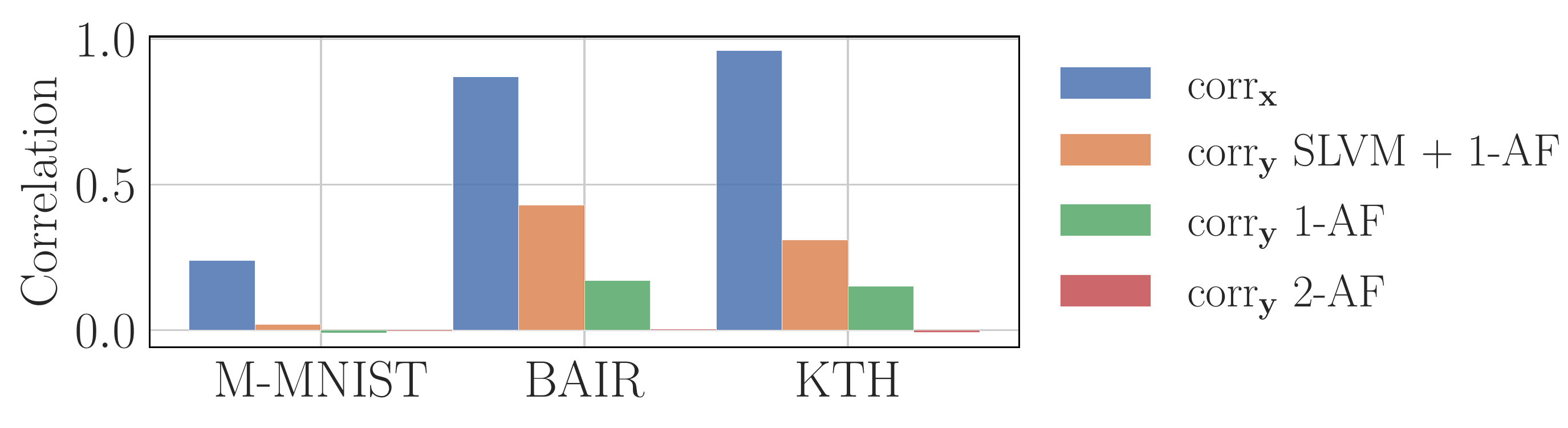}
    \caption{}
    \label{fig: corr bar plot}
    \end{subfigure}
    ~
    \begin{subfigure}[t]{0.25\textwidth}
    \centering
    \includegraphics[width=\textwidth]{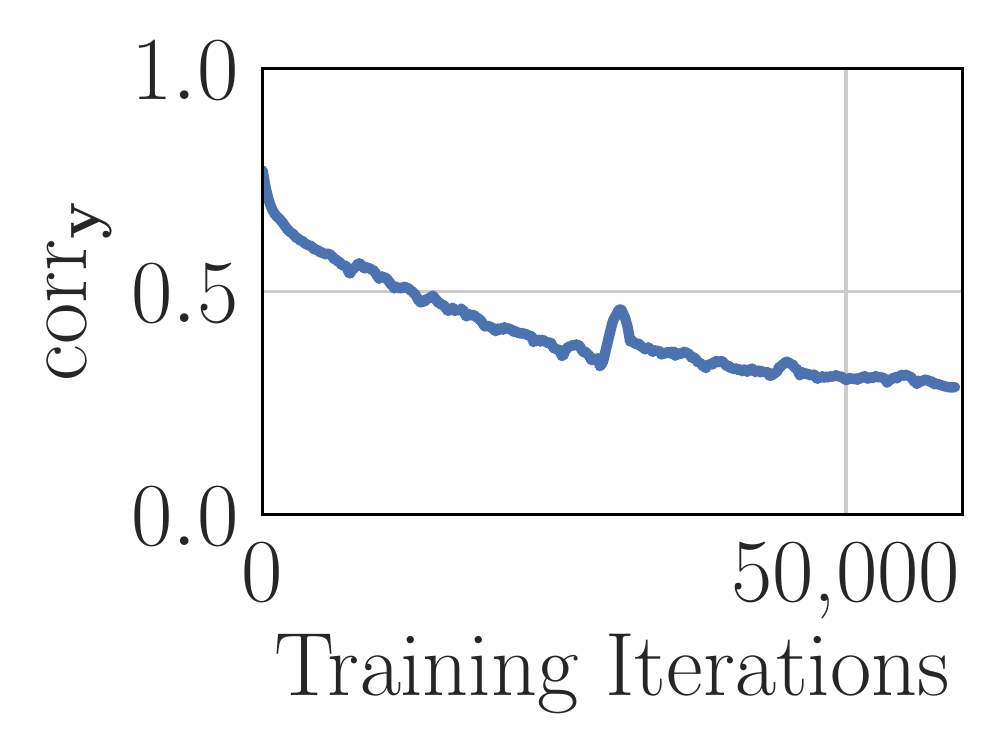}
    \caption{}
    \label{fig: noise corr kth}
    \end{subfigure}%
    \caption{\textbf{Decreased Temporal Correlation}. \textbf{(a)} Affine autoregressive flows result in sequences, $\mathbf{y}_{1:T}$, with decreased temporal correlation, $\textrm{corr}_\mathbf{y}$, as compared with that of the original data, $\textrm{corr}_\mathbf{x}$. The presence of a more powerful base distribution (SLVM) reduces the need for decorrelation. Additional flow transforms further decrease correlation (\textit{note}: $|\textrm{corr}_\mathbf{y}| < 0.01$ for 2-AF). \textbf{(b)} For SLVM + 1-AF, $\textrm{corr}_\mathbf{y}$ decreases during training on KTH Actions.}
\end{figure}

\subsection{Experimental Setup}
\label{sec:exp_setup}

We empirically evaluate the improvements to downstream dynamics modeling from temporal pre-processing via autoregressive flows. For data space modeling, we compare four model classes: \textbf{1)} standalone affine autoregressive flows with one (1-AF) and \textbf{2)} two (2-AF) transforms, \textbf{3)} a sequential latent variable model (SLVM), and \textbf{4)} SLVM with flow-based pre-processing (SLVM + 1-AF). As we are not proposing a specific architecture, but rather a general modeling technique, the SLVM architecture is representative of recurrent convolutional video models with a single latent level \citep{denton2018stochastic,ha2018recurrent,hafner2019learning}. Flows are implemented with convolutional networks, taking in a fixed window of previous frames (Fig.~\ref{fig: flow architecture}). These models allow us to evaluate the benefits of temporal pre-processing (SLVM vs. SLVM + 1-AF) and the benefits of more expressive higher-level dynamics models (2-AF vs. SLVM + 1-AF).

To evaluate latent dynamics modeling with flows, we use the \texttt{tensor2tensor} library \citep{tensor2tensor} to compare \textbf{1)} VideoFlow\footnote{We used a smaller version of the original model architecture, with half of the flow depth, due to GPU memory constraints.} and \textbf{2)} the same model with affine autoregressive flow latent dynamics (VideoFlow + AF). VideoFlow is significantly larger ($3\times$ more parameters) than the one-level SLVM, allowing us to evaluate whether autoregressive flows are beneficial in this high-capacity regime.

To enable a fairer comparison in our experiments, models with autoregressive flow dynamics have comparable or fewer parameters than baseline counterparts. We note that autoregressive dynamics adds only a constant computational cost per time-step, and this computation can be parallelized for training and evaluation. Full architecture, training, and analysis details can be found in the Appendix. Finally, as noted by  \cite{kumar2019videoflow}, many previous works do not train SLVMs with proper log-likelihood objectives. Our SLVM results are consistent with previously reported log-likelihood values \citep{marino2018general} for the Stochastic Video Generation model \citep{denton2018stochastic} trained with a log-likelihood bound objective.

\begin{figure}[t!]
    \centering
    \begin{subfigure}[t]{0.043\textwidth}
        \centering
        \includegraphics[width=\textwidth]{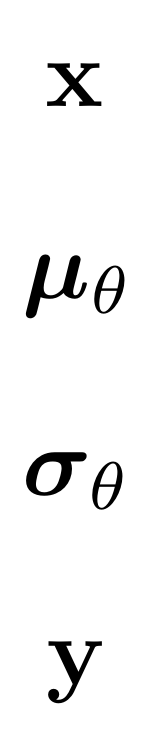}
    \end{subfigure}%
    ~ 
    \begin{subfigure}[t]{0.44\textwidth}
        \centering
        \includegraphics[width=\textwidth]{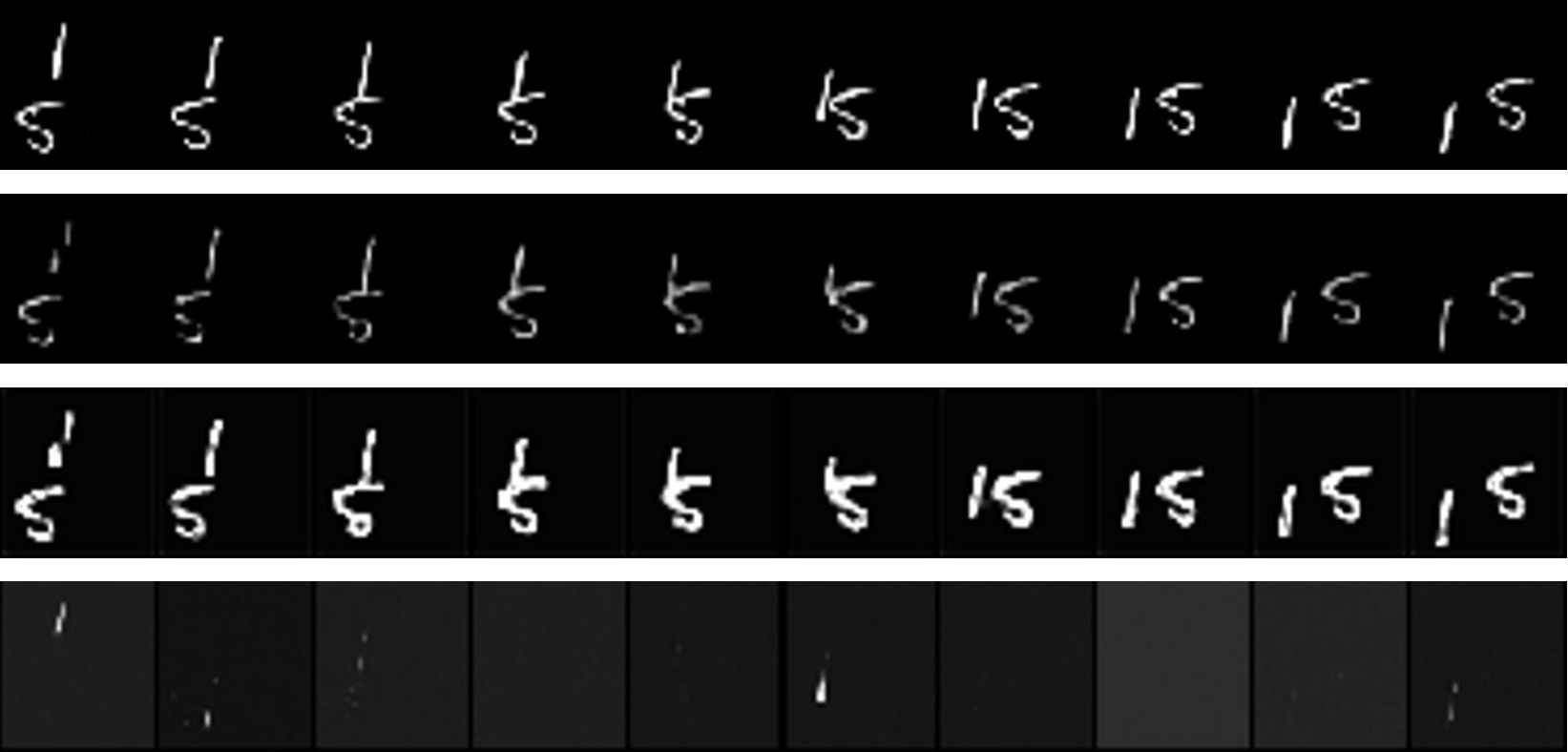}
        \label{fig: latent + flow vis mnist}
    \end{subfigure}%
    ~ 
    \begin{subfigure}[t]{0.44\textwidth}
        \centering
        \includegraphics[width=\textwidth]{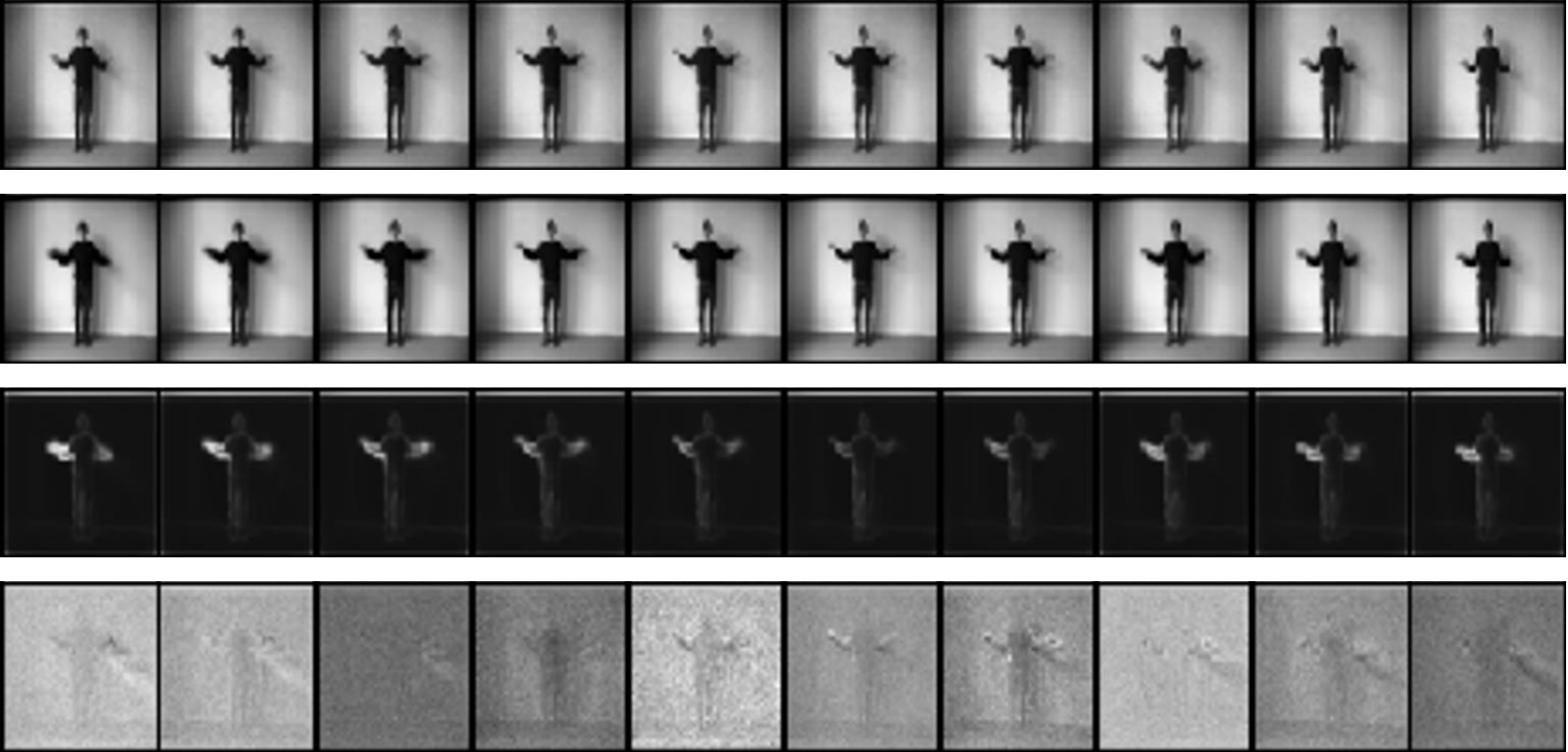}
        \label{fig: latent + flow vis kth}
    \end{subfigure}
    \caption{\textbf{Flow Visualization} for SLVM + 1-AF on Moving MNIST (left) and KTH Actions (right).}
    \label{fig: flow visualization}
\end{figure}

\begin{figure}[t!]
    \centering
    \begin{subfigure}[h]{0.85\textwidth}
        \centering
        \includegraphics[width=\textwidth]{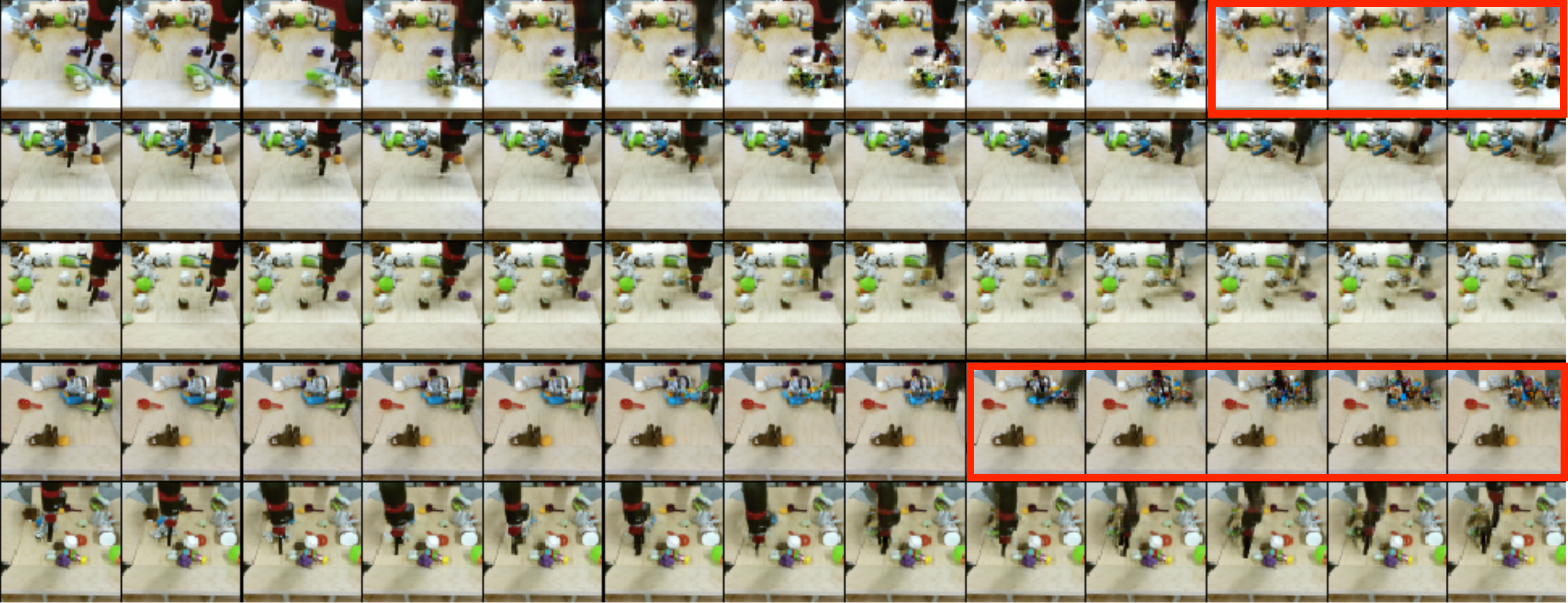}
        \label{fig: videoflow gen subfig}
        \vspace{-10pt}
        \caption{VideoFlow}
        \vspace{10pt}
    \end{subfigure}%

    \begin{subfigure}[h]{0.85\textwidth}
        \centering
        \includegraphics[width=\textwidth]{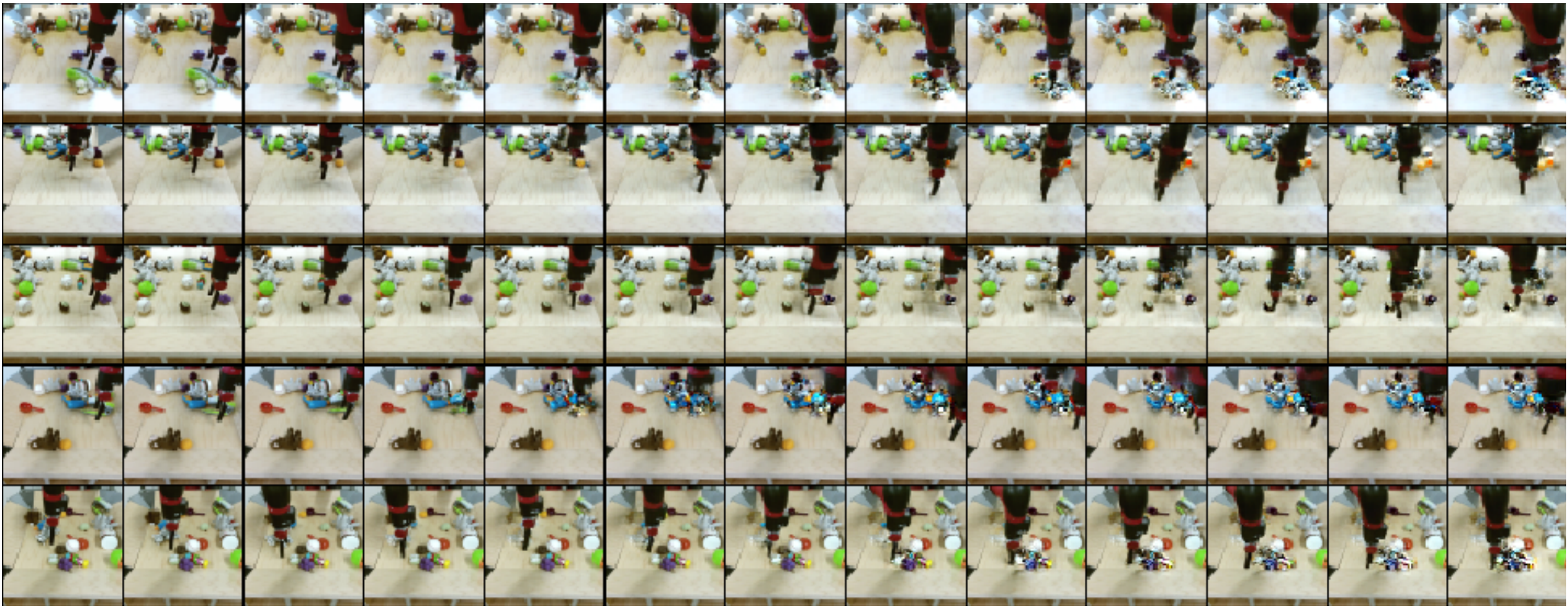}
        \label{fig: ar videoflow gen}
        \vspace{-10pt}
        \caption{VideoFlow + AF}
        \vspace{10pt}
    \end{subfigure}
    \caption{\textbf{Improved Generated Samples}. Random samples generated from (\textbf{a}) VideoFlow and (\textbf{b}) VideoFlow + AF, each conditioned on the first $3$ frames. Using AF produces more coherent samples. The robot arm blurs for VideoFlow in samples $1$ and $4$ (red boxes), but does not blur for VideoFlow + AF.}
    \label{fig: videoflow gen}
\end{figure}

\subsection{Analyses}
\label{sec:exp_qualitative}

\paragraph{Visualization} In Figure~\ref{fig:data vs noise}, we visualize the pre-processing procedure for SLVM + 1-AF on BAIR Robot Pushing. The plots show the RGB values for a pixel before (top) and after (bottom) the transform. The noise sequence is nearly zero throughout, despite large changes in the pixel value. We also see that the noise sequence (center, lower) is invariant to the static background, capturing the moving robotic arm. At some time steps (e.g. fourth frame), the autoregressive flow incorrectly predicts the next frame, however, the higher-level SLVM compensates for this prediction error.

We also visualize each component of the flow. Figure~\ref{fig: flow architecture} illustrates this for SLVM + 1-AF on an input from BAIR Robot Pushing. We see that $\bm{\mu}_\theta$ captures the static background, while $\bm{\sigma}_\theta$ highlights regions of uncertainty. In Figure~\ref{fig: flow visualization} and the Appendix, we present visualizations on full sequences, where we see that different models remove varying degrees of temporal structure.

\paragraph{Temporal Redundancy Reduction} To quantify temporal redundancy reduction, we evaluate the empirical correlation (linear dependence) between frames, denoted as corr, for the data and noise variables. We evaluate $\textrm{corr}_{\mathbf{x}}$ and $\textrm{corr}_{\mathbf{y}}$ for 1-AF, 2-AF, and SLVM + 1-AF. The results are shown in Figure~\ref{fig: corr bar plot}. In Figure~\ref{fig: noise corr kth}, we plot $\textrm{corr}_{\mathbf{y}}$ for SLVM + 1-AF during training on KTH Actions. Flows decrease temporal correlation, with additional transforms yielding further decorrelation. Base distributions without temporal structure (1-AF) yield comparatively more decorrelation. Temporal redundancy is progressively removed throughout training. Note that 2-AF almost completely removes temporal correlations ($|\textrm{corr}_\mathbf{y}| < 0.01$). However, note that this only quantifies \textit{linear} dependencies, and more complex non-linear dependencies may require the use of higher-level dynamics models, as shown through quantitative comparisons.

\paragraph{Performance Comparison} Table~\ref{table:elbo_comparison} reports average test negative log-likelihood results on video datasets. Standalone flow-based models perform surprisingly well. Increasing flow depth from 1-AF to 2-AF generally results in improvement. SLVM + 1-AF outperforms the baseline SLVM despite having \textit{fewer} parameters. As another baseline, we also consider modeling frame differences, $\Delta \mathbf{x} \equiv \mathbf{x}_t - \mathbf{x}_{t-1}$, with SLVM, which can be seen as a special case of 1-AF with $\bm{\mu}_\theta = \mathbf{x}_{t-1}$ and $\bm{\sigma}_\theta = \mathbf{1}$. On BAIR and KTH Actions, datasets with significant temporal redundancy (Fig.~\ref{fig: corr bar plot}), this technique improves performance over SLVM. However, on Moving MNIST, modeling $\Delta \mathbf{x}$ actually decreases performance, presumably by creating more complex spatial patterns. In all cases, the learned temporal transform, SLVM + 1-AF, outperforms this hard-coded transform, SLVM + $\Delta \mathbf{x}$. Finally, incorporating autoregressive flows into VideoFlow results in a modest but noticeable improvement, demonstrating that removing spatial dependencies, through VideoFlow, and temporal dependencies, through autoregressive flows, are complementary techniques.

\begin{table}[t]
\caption{{\bf Quantitative Comparison.} Average test negative log-likelihood (lower is better) in \textit{nats per dimension} for Moving MNIST, BAIR Robot Pushing, and KTH Actions.}
\label{table:elbo_comparison}
\begin{tabular}{lrrr}
\toprule
 & M-MNIST & BAIR & KTH  \\
\midrule
1-AF      & $2.15$  &  $3.05$ & $3.34$ \\
2-AF      & $2.13$  &  $2.90$ & $3.35$ \\
\bottomrule
SLVM      & $\leq 1.92$  &  $\leq 3.57$ & $\leq 4.63$ \\
SLVM + $\Delta \mathbf{x}$   & $\leq 2.45$  & $\leq 3.07$  & $\leq 2.49$ \\
SLVM + 1-AF      & $\leq \mathbf{1.86}$  &  $\leq \mathbf{2.35}$ & $\leq \mathbf{2.39}$ \\
\bottomrule
VideoFlow & -- & $1.53$ & -- \\
VideoFlow + AF & -- &  $\mathbf{1.50}$ & -- \\
\bottomrule
\end{tabular}
\end{table}
\label{sec:exp_quantitative}

\begin{figure}[t!]
    \centering
    \begin{subfigure}[t]{0.32\textwidth}
    \centering
    \includegraphics[width=\textwidth]{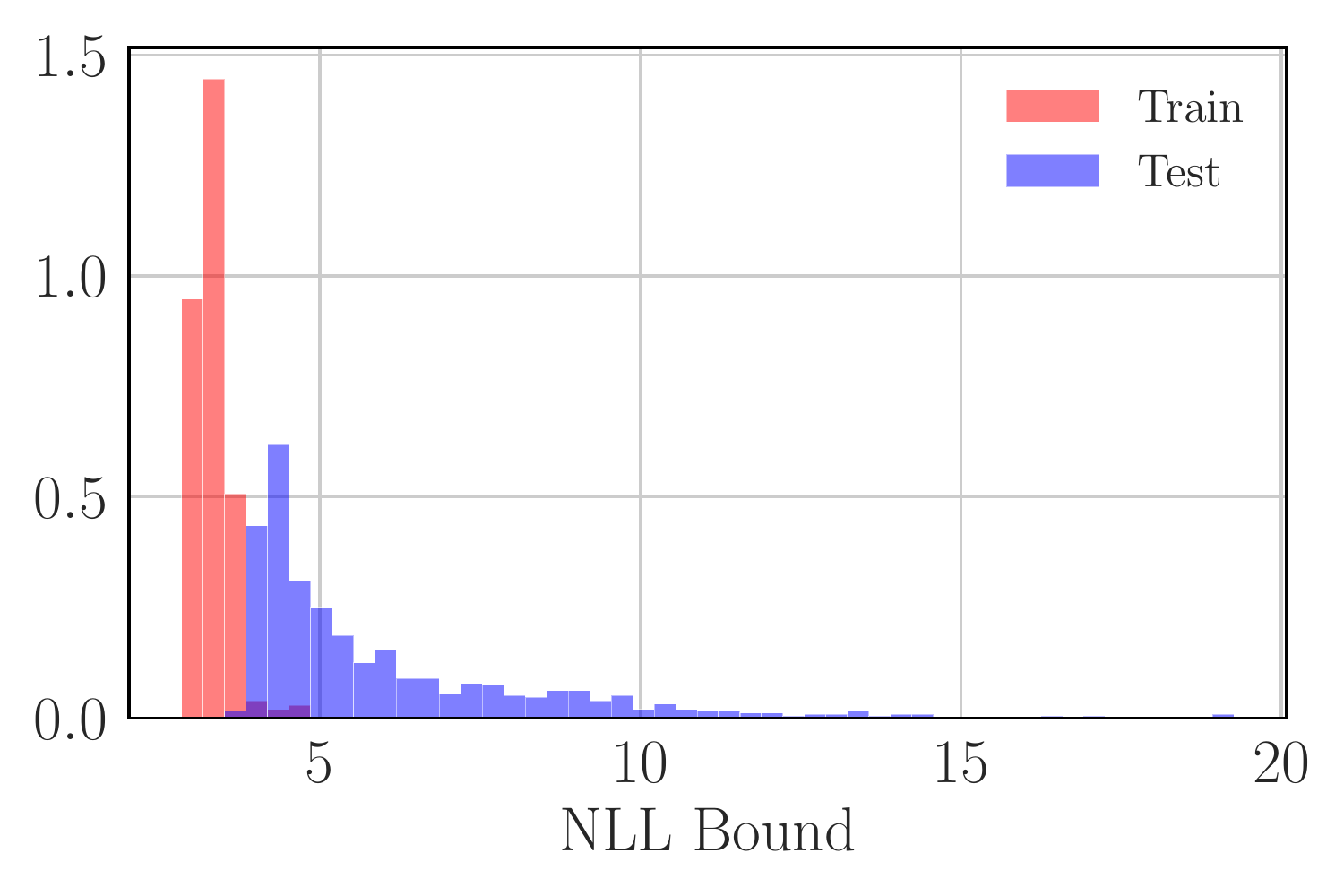}
    \caption{SLVM}
    \label{fig: slvm nll hist}
    \end{subfigure}
    ~
    \begin{subfigure}[t]{0.32\textwidth}
    \centering
    \includegraphics[width=\textwidth]{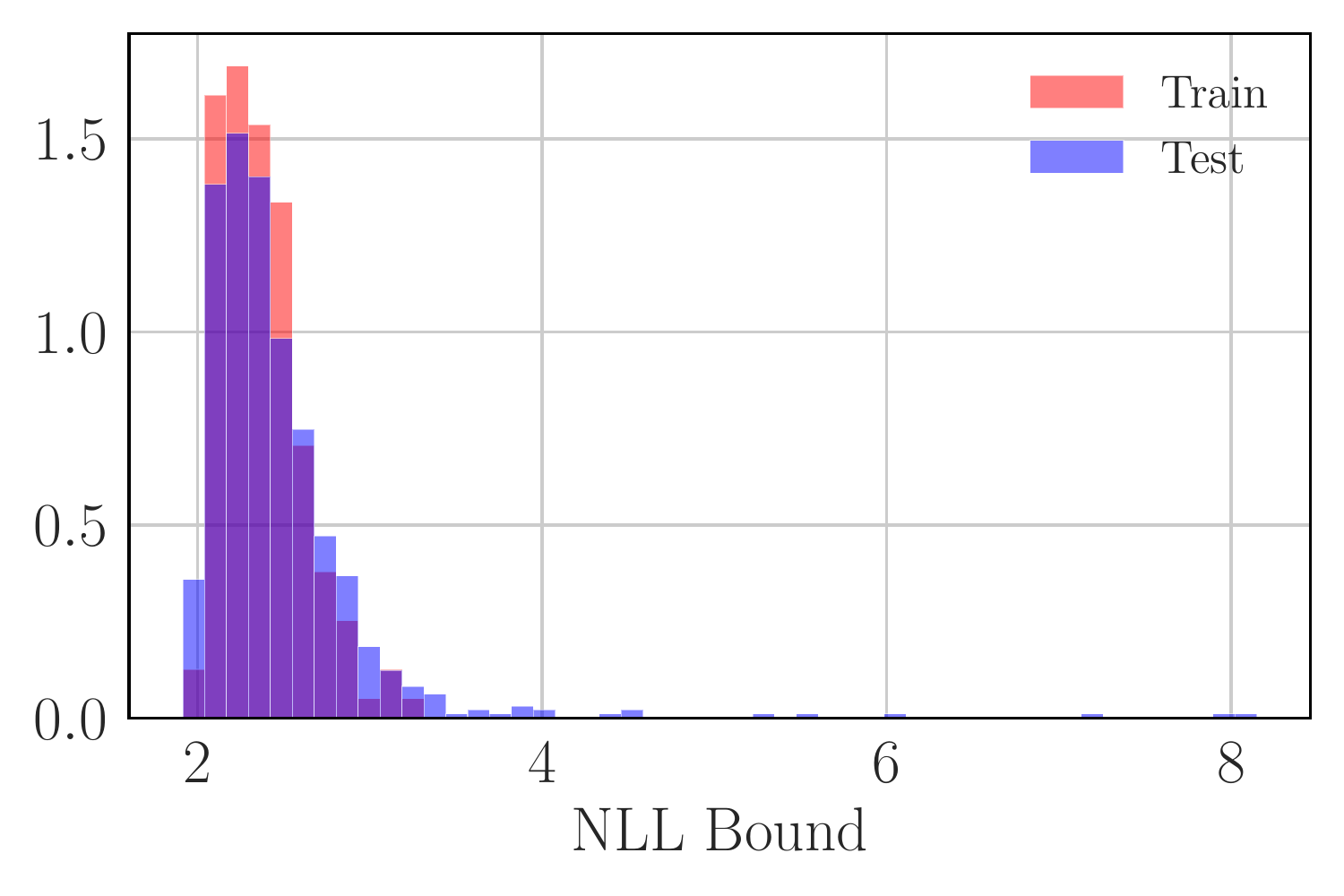}
    \caption{SLVM + 1-AF}
    \label{fig: slvm + af nll hist}
    \end{subfigure}%
    ~
    \begin{subfigure}[t]{0.27\textwidth}
    \centering
    \includegraphics[width=\textwidth]{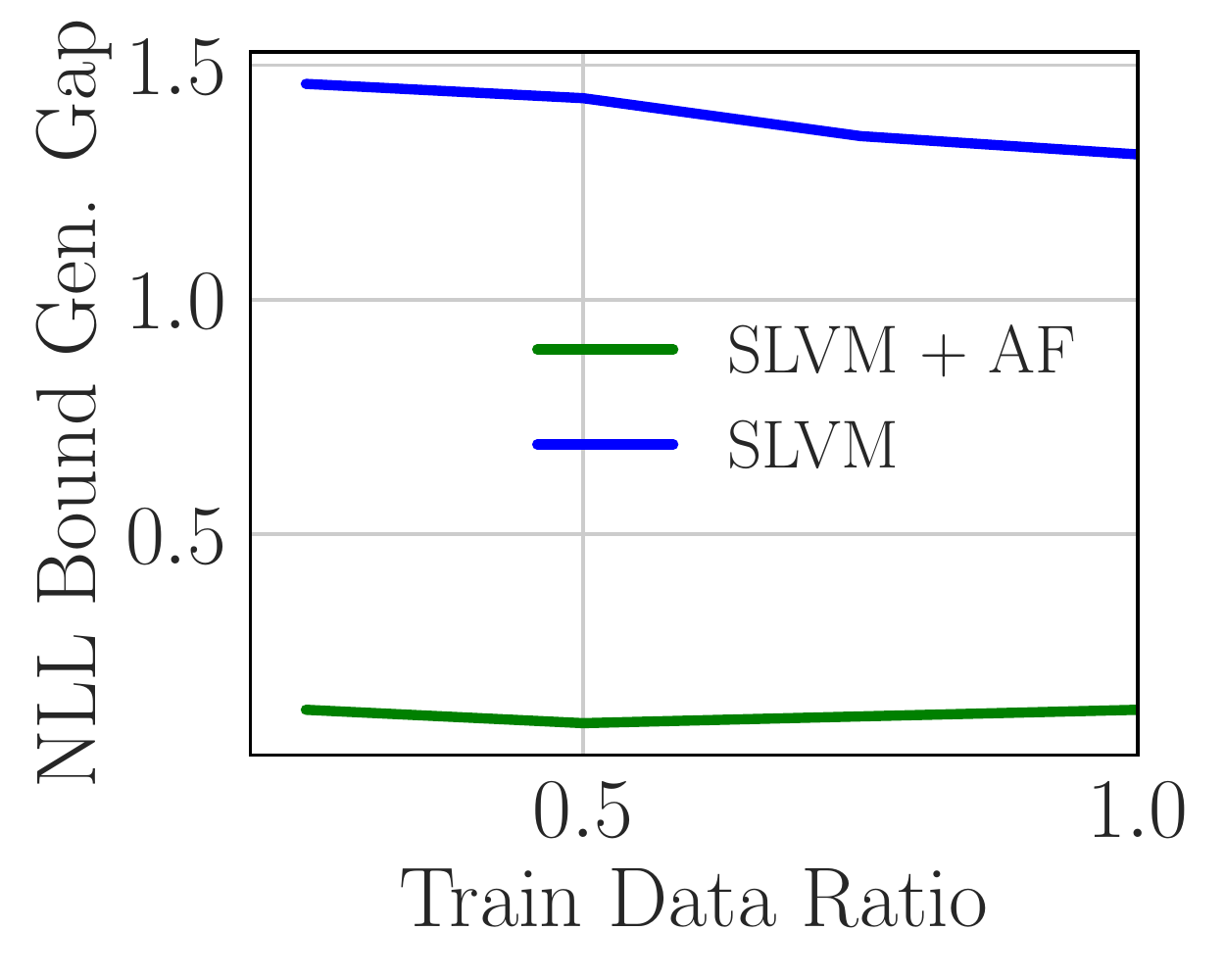}
    \caption{}
    \label{fig: gen gap}
    \end{subfigure}%
    \caption{\textbf{Improved Generalization}. The low-level reference frame improves generalization to unseen sequences. Train and test negative log-likelihood bound histograms for \textbf{(a)} SLVM and \textbf{(b)} SLVM + 1-AF on KTH Actions. \textbf{(c)} The generalization gap for SLVM + 1-AF remains small for varying amounts of KTH training data, while it becomes worse in the low-data regime for SLVM.}
    \label{fig: nll hists}
\end{figure}

\paragraph{Results on Non-Video Sequence Dataset}
In Table \ref{tab: non-video results}, we report negative log-density results on non-video data in \textit{nats per time step}. Note that log-densities can be positive or negative. Again, we see that 2-AF consistently outperforms 1-AF, which are typically on-par or better than SLVM. However, SLVM + 1-AF outperforms all other model classes, achieving the lowest (best) log-densities across all datasets. With non-video data, we see that using the special case of modeling temporal differences (SLVM + $\Delta \mathbf{x}$), performance is actually slightly \textit{worse} than that of SLVM on all datasets. This, again, highlights the importance of using a learned pre-processing transform in comparison with hard-coded temporal differences.

\begin{table}[th]
\caption{\textbf{Non-Video Quantitative Comparison.} Average test log-likelihood (lower is better) in \textit{nats per time step} on various non-video datasets.}
\label{tab: non-video results}
\centering
\begin{tabular}{l  c  c  c}
\toprule
 & \texttt{activity\_rec} & \texttt{smartphone\_sensor} &   \texttt{facial\_exp} \\
\hline
 1-AF & $2.71$  & $-7.46$ &  $-241$ \\
 2-AF & $2.06$   & $-8.53$  & $-259$ \\
\hline
 SLVM & $\leq 2.77$  & $\leq -5.21$  & $\leq -164$ \\
 SLVM + $\Delta \mathbf{x}$   & $\leq 5.61$  & $\leq -4.02$  & $\leq -154$ \\
 SLVM + 1-AF & $\mathbf{\leq 1.46}$ & $\mathbf{\leq -9.82}$ & $\mathbf{\leq -306}$ \\
\bottomrule
\end{tabular}
\end{table}

\paragraph{Improved Samples} The quantitative improvement over VideoFlow is less dramatic, as this is already a high-capacity model. However, qualitatively, we observe that incorporating autoregressive flow dynamics improves sample quality (Figure~\ref{fig: videoflow gen}). In these randomly selected samples, the robot arm occasionally becomes blurry for VideoFlow (red boxes) but remains clear for VideoFlow + AF.

\paragraph{Improved Generalization} Our temporal normalization technique also improves generalization to unseen examples, a key benefit of normalization schemes, e.g., batch norm \citep{ioffe2015batch}. Intuitively, higher-level dynamics are often preserved, whereas lower-level appearance is not. This is apparent on KTH Actions, which contains a substantial degree of train-test mismatch, due to different identities and activities. NLL histograms on KTH are shown in Figure~\ref{fig: nll hists}, with greater overlap for SLVM + 1-AF. We also train SLVM and SLVM + 1-AF on subsets of KTH Actions. In Figure~\ref{fig: gen gap}, we see that autoregressive flows enable generalization in the low-data regime, whereas SLVM becomes worse.

\section{Conclusion}
\label{sec: conclusion}

We have presented a technique for improving sequence modeling using autoregressive flows. Learning a frame of reference, parameterized by autoregressive transforms, reduces temporal redundancy in input sequences, simplifying dynamics. Thus, rather than expanding the model, we can simplify the input to meet the capacity of the model. This approach is distinct from previous works with normalizing flows on sequences, yet contains connections to classical modeling and compression. We hope these connections lead to further insights and applications. Finally, we have analyzed and empirically shown how autoregressive pre-processing in both the data and latent spaces can improve sequence modeling and lead to improved sample quality and generalization.

The underlying assumption behind using autoregressive flows for sequence modeling is that sequences contain smooth or predictable temporal dependencies, with more complex, higher-level dependencies as well. In both video and non-video data, we have seen improvements from combining sequential latent variable models with autoregressive flows, suggesting that such assumptions are generally reasonable. Using affine autoregressive flows restricts our approach to sequences of continuous data, but future work could investigate discrete data, such as natural language. Likewise, we assume regularly sampled sequences (i.e., a constant frequency), however, future work could also investigate irregularly sampled event data.

\nocite{schmidt2019autoregressive}
\nocite{lombardo2019deep}

\bibliography{main}
\bibliographystyle{plainnat}

\newpage
\appendix

\section{Lower Bound Derivation}
\label{sec: lower bound derivation}

Consider the model defined in Section 3.3, with the conditional likelihood parameterized with autoregressive flows. That is, we parameterize
\begin{equation}
    \mathbf{x}_t = \bm{\mu}_\theta (\mathbf{x}_{<t}) + \bm{\sigma}_\theta (\mathbf{x}_{<t}) \odot \mathbf{y}_t
\end{equation}
yielding
\begin{equation}
    p_\theta (\mathbf{x}_t | \mathbf{x}_{<t}, \mathbf{z}_{\leq t}) = p_\theta (\mathbf{y}_t | \mathbf{y}_{<t}, \mathbf{z}_{\leq t}) \left| \det \left( \frac{\partial \mathbf{x}_t}{\partial \mathbf{y}_t} \right) \right|^{-1}.
\end{equation}
The joint distribution over all time steps is then given as
\begin{align}
    p_\theta (\mathbf{x}_{1:T}, \mathbf{z}_{1:T}) & = \prod_{t=1}^T p_\theta (\mathbf{x}_t | \mathbf{x}_{<t}, \mathbf{z}_{\leq t}) p_\theta (\mathbf{z}_t | \mathbf{x}_{<t}, \mathbf{z}_{<t}) \\
    & = \prod_{t=1}^T p_\theta (\mathbf{y}_t | \mathbf{y}_{<t}, \mathbf{z}_{\leq t}) \left| \det \left( \frac{\partial \mathbf{x}_t}{\partial \mathbf{y}_t} \right) \right|^{-1} p_\theta (\mathbf{z}_t | \mathbf{x}_{<t}, \mathbf{z}_{<t}).
\end{align}
To perform variational inference, we consider a filtering approximate posterior of the form
\begin{equation}
    q (\mathbf{z}_{1:T} | \mathbf{x}_{1:T}) = \prod_{t=1}^T q (\mathbf{z}_t | \mathbf{x}_{\leq t}, \mathbf{z}_{<t}).
\end{equation}
We can then plug these expressions into the evidence lower bound:
\begin{align}
    \mathcal{L} & \equiv \mathbb{E}_{q (\mathbf{z}_{1:T} | \mathbf{x}_{1:T})} \left[ \log p_\theta (\mathbf{x}_{1:T}, \mathbf{z}_{1:T}) - \log q (\mathbf{z}_{1:T} | \mathbf{x}_{1:T}) \right] \\
    & = \mathbb{E}_{q (\mathbf{z}_{1:T} | \mathbf{x}_{1:T})} \Bigg[ \log \left( \prod_{t=1}^T p_\theta (\mathbf{y}_t | \mathbf{y}_{<t}, \mathbf{z}_{\leq t}) \left| \det \left( \frac{\partial \mathbf{x}_t}{\partial \mathbf{y}_t} \right) \right|^{-1} p_\theta (\mathbf{z}_t | \mathbf{x}_{<t}, \mathbf{z}_{<t}) \right) \nonumber \\
    & \ \ \ \ \ \ \ \ \ \ \ \ \ \ \ \ \ \ \ \ \ \ \ \ \ \ \ \ - \log \left( \prod_{t=1}^T q (\mathbf{z}_t | \mathbf{x}_{\leq t}, \mathbf{z}_{<t}) \right) \Bigg]  \\
    & = \mathbb{E}_{q (\mathbf{z}_{1:T} | \mathbf{x}_{1:T})} \Bigg[ \sum_{t=1}^T \log p_\theta (\mathbf{y}_t | \mathbf{y}_{<t}, \mathbf{z}_{\leq t}) - \log \frac{q (\mathbf{z}_t | \mathbf{x}_{\leq t}, \mathbf{z}_{<t})}{p_\theta (\mathbf{z}_t | \mathbf{x}_{<t}, \mathbf{z}_{<t})} - \log \left| \det \left( \frac{\partial \mathbf{x}_t}{\partial \mathbf{y}_t} \right) \right| \Bigg] .
\end{align}
Finally, in the filtering setting, we can rewrite the expectation, bringing it inside of the sum (see \cite{gemici2017generative,marino2018general}):
\begin{equation}
    \mathcal{L} = \sum_{t=1}^T \mathbb{E}_{q (\mathbf{z}_{\leq t} | \mathbf{x}_{\leq t})} \Bigg[ \log p_\theta (\mathbf{y}_t | \mathbf{y}_{<t}, \mathbf{z}_{\leq t}) - \log \frac{q (\mathbf{z}_t | \mathbf{x}_{\leq t}, \mathbf{z}_{<t})}{p_\theta (\mathbf{z}_t | \mathbf{x}_{<t}, \mathbf{z}_{<t})} - \log \left| \det \left( \frac{\partial \mathbf{x}_t}{\partial \mathbf{y}_t} \right) \right| \Bigg] .
\end{equation}
Because there exists a one-to-one mapping between $\mathbf{x}_{1:T}$ and $\mathbf{y}_{1:T}$, we can equivalently condition the approximate posterior and the prior on $\mathbf{y}$, i.e.
\begin{equation}
    \mathcal{L} = \sum_{t=1}^T \mathbb{E}_{q (\mathbf{z}_{\leq t} | \mathbf{y}_{\leq t})} \Bigg[ \log p_\theta (\mathbf{y}_t | \mathbf{y}_{<t}, \mathbf{z}_{\leq t}) - \log \frac{q (\mathbf{z}_t | \mathbf{y}_{\leq t}, \mathbf{z}_{<t})}{p_\theta (\mathbf{z}_t | \mathbf{y}_{<t}, \mathbf{z}_{<t})} - \log \left| \det \left( \frac{\partial \mathbf{x}_t}{\partial \mathbf{y}_t} \right) \right| \Bigg] .
\end{equation}
\newpage
\section{Experiment Details}
\label{sec: experiment details}

\subsection{Flow Architecture}

The affine autoregressive flow architecture is shown in Figure~\ref{fig: model diagrams}. The shift and scale of the affine transform are conditioned on three previous inputs. For each flow, we first apply $4$ convolutional layers with kernel size $(3,3)$, stride $1$, and padding $1$ on each conditioned observation, preserving the input shape. The outputs are concatenated along the channel dimension and go through another $4$ convolutional layers with kernel size $(3,3)$, stride $1$, and padding $1$. Finally, separate convolutional layers with the same kernel size, stride, and padding are used to output shift and log-scale. We use ReLU non-linearities for all convolutional layers.

\subsection{Sequential Latent Variable Model Architecture}

\begin{figure}[t!]
    \centering
    \begin{subfigure}[t]{0.22\textwidth}
        \centering
        \includegraphics[width=\textwidth]{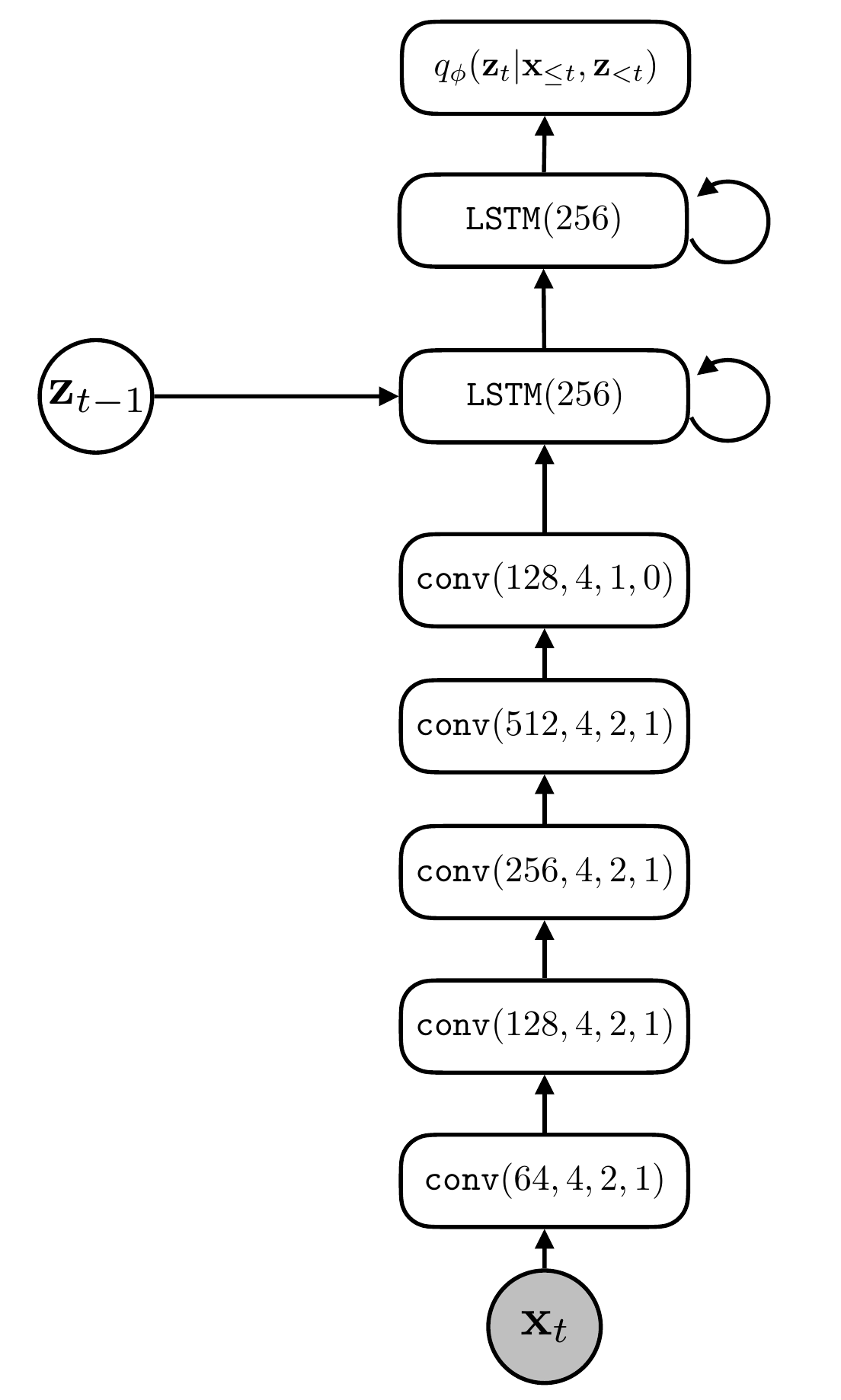}
        \caption{}
        \label{fig: encoder arch}
    \end{subfigure}%
    ~ 
    \begin{subfigure}[t]{0.45\textwidth}
        \centering
        \includegraphics[width=\textwidth]{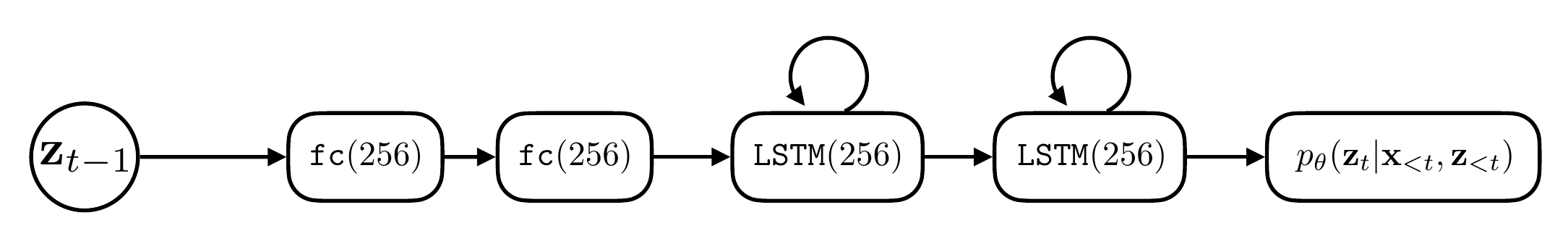}
        \caption{}
        \label{fig: prior arch}
    \end{subfigure}
    ~ 
    \begin{subfigure}[t]{0.16\textwidth}
        \centering
        \includegraphics[width=\textwidth]{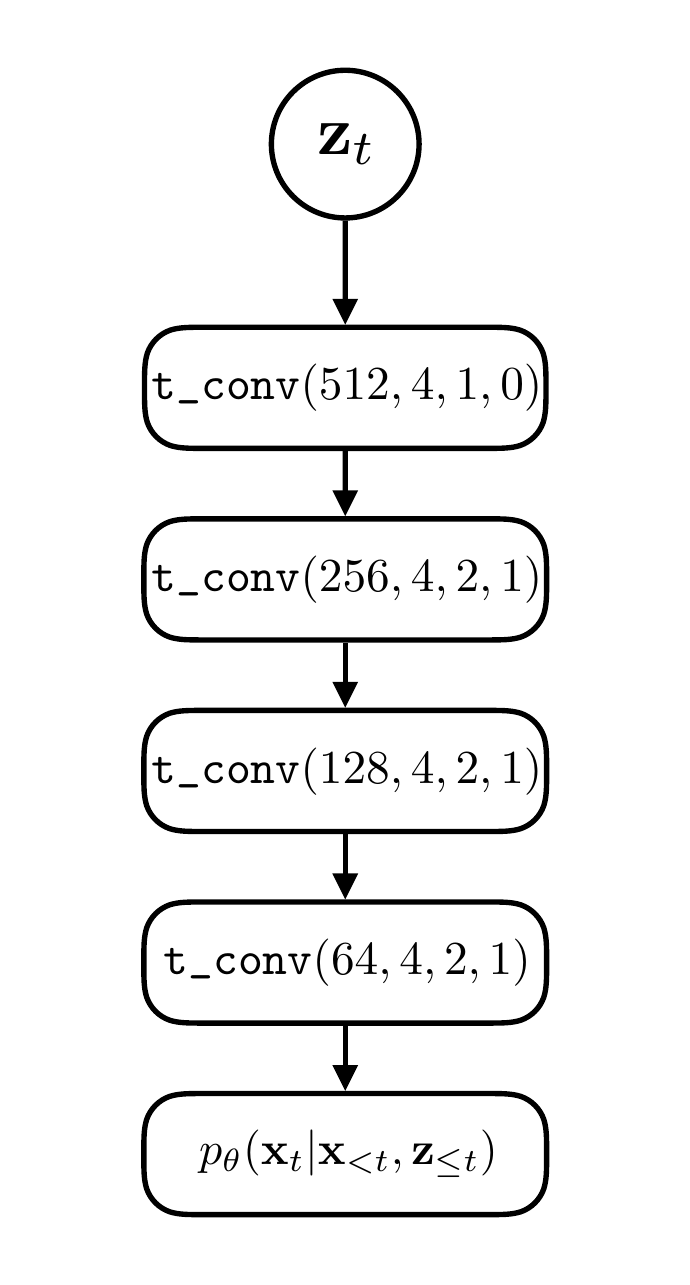}
        \caption{}
        \label{fig: decoder arch}
    \end{subfigure}
    
    \begin{subfigure}[t]{0.22\textwidth}
        \centering
        \includegraphics[width=\textwidth]{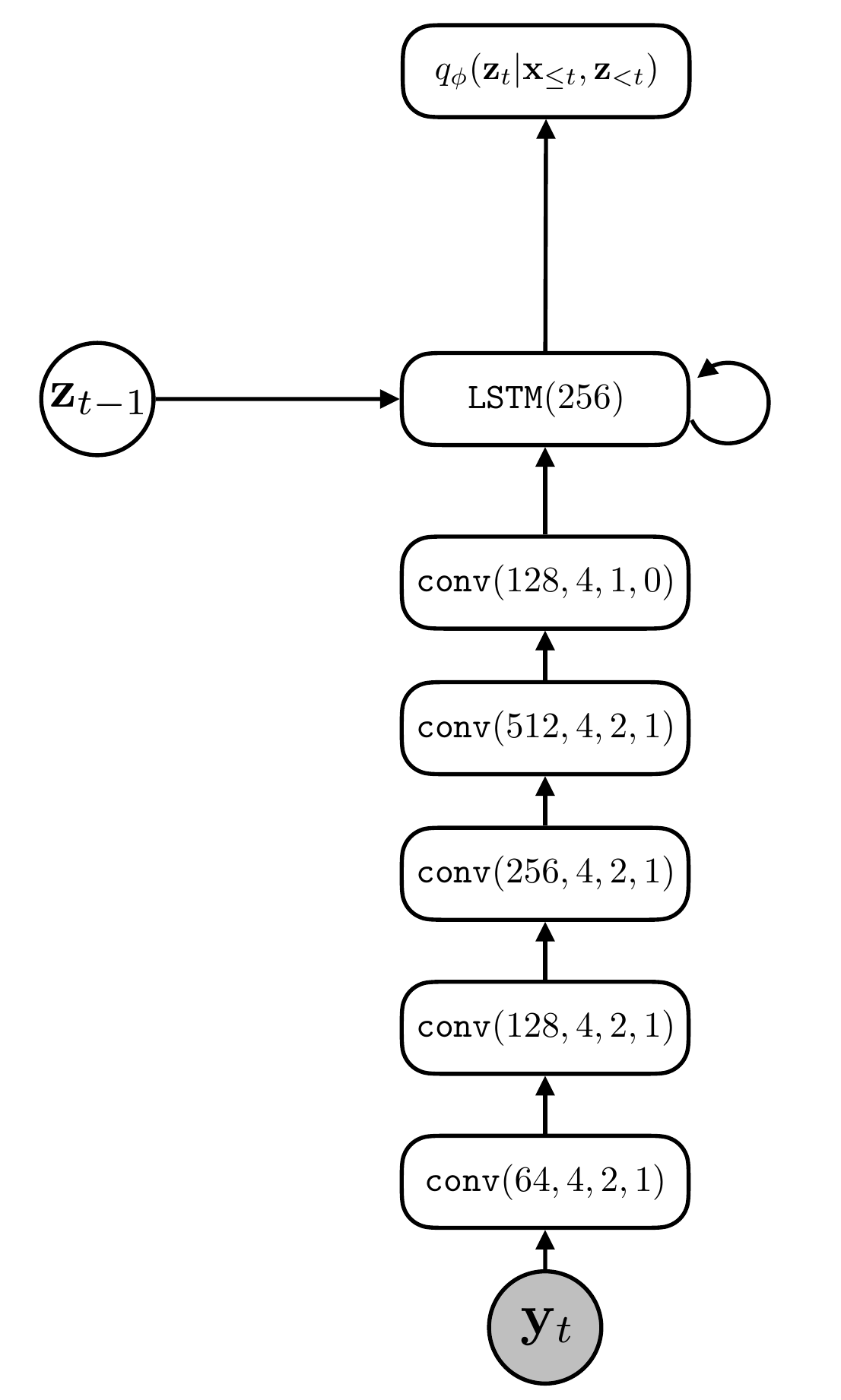}
        \caption{}
        \label{fig: encoder arch}
    \end{subfigure}%
    ~ 
    \begin{subfigure}[t]{0.45\textwidth}
        \centering
        \includegraphics[width=\textwidth]{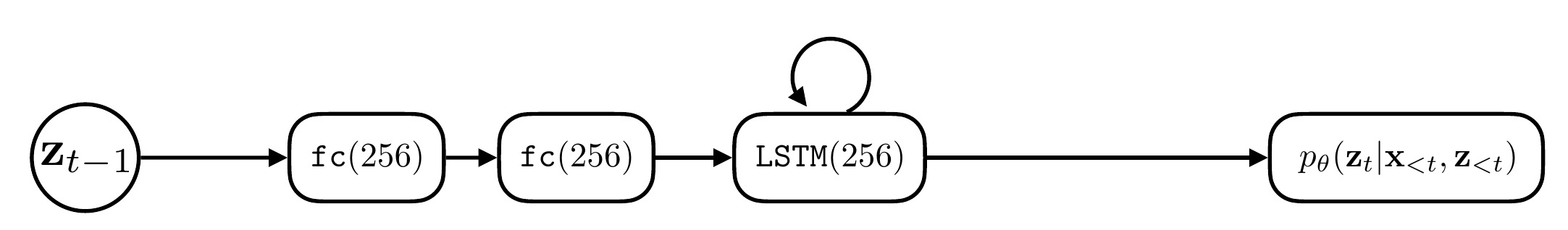}
        \caption{}
        \label{fig: prior arch}
    \end{subfigure}    \caption{\textbf{SLVM Architecture}. Diagrams are shown for the (\textbf{a}) approximate posterior, (\textbf{b}) prior, and (\textbf{c}) conditional likelihood of the sequential latent variable model (SLVM). In (\textbf{d}) and (\textbf{e}) we show the approximate posterior and prior used with SLVM + AF, respectively. The conditional likelihood is the same architecture in both setups. Note: for SLVM + AF, we input $\mathbf{y}_t$ into the approximate posterior encoder, rather than $\mathbf{x}_t$. \texttt{conv} denotes a convolutional layer, \texttt{LSTM} denotes a long short-term memory layer, \texttt{fc} denotes a fully-connected layer, and \texttt{t\_conv} denotes a transposed convolutional layer. For \texttt{conv} and \texttt{t\_conv} layers, the numbers in parentheses respectively denote the number of filters, filter size, stride, and padding of the layer. For \texttt{fc} and \texttt{LSTM} layers, the number in parentheses denotes the number of units. SLVM contains one additional LSTM layer in both the approximate posterior and conditional prior. }
    \label{fig: model diagrams}
\end{figure}

For sequential latent variable models, we use a DC-GAN \citep{radford2015unsupervised} encoder architecture (Figure~\ref{fig: encoder arch}), with $4$ convolutional layers of kernel size $(4,4)$, stride $2$, and padding $1$ followed by another convolutional layer of kernel size $(4,4)$, stride $1$, and no padding. The encoding is sent to one or two LSTM layers \citep{hochreiter1997long} followed by separate linear layers to output the mean and log-variance for $q_\phi (\mathbf{z}_t | \mathbf{x}_{\leq t}, \mathbf{z}_{<t})$. We note that for SLVM, we input $\mathbf{x}_t$ into the encoder, whereas for SLVM + AF, we input $\mathbf{y}_t$. The architecture for the conditional prior, $p_\theta (\mathbf{z}_t | \mathbf{x}_{<t}, \mathbf{z}_{<t})$, shown in Figure~\ref{fig: prior arch}, contains two fully-connected layers,  which take the previous latent variable as input, followed by one or two LSTM layers, and separate linear layers to output the mean and log-variance. The decoder architecture, shown in Figure~\ref{fig: decoder arch}, mirrors the encoder architecture, using transposed convolutions. In SLVM, we use two LSTM layers for modeling the conditional prior and approximate posterior distributions, while in SLVM + 1-AF, we use a single LSTM layer for each. We use leaky ReLU non-linearities for the encoder and decoder architectures and ReLU non-linearities in the conditional prior architecture.

\subsection{VideoFlow Architecture}

For VideoFlow experiments, we use the official code provided by \cite{kumar2019videoflow} in the \texttt{tensor2tensor} repository \citep{tensor2tensor}. Due to memory and computational constraints, we use a smaller version of the model architecture used by \cite{kumar2019videoflow} for the BAIR Robot Pushing dataset. We change \texttt{depth} from $24$ to $12$ and \texttt{latent\_encoder\_width} from $256$ to $128$. This reduces the number of parameters from roughly $67$ million to roughly $32$ million. VideoFlow contains a hierarchy of latent variables, with the latent variable at level $l$ at time $t$ denoted as $\mathbf{z}_t^{(l)}$. The prior on this latent variable is denoted as $p_\theta (\mathbf{z}_t^{(l)} | \mathbf{z}_{<t}^{(l)}, \mathbf{z}_t^{(>l)}) = \mathcal{N} (\mathbf{z}_t^{(l)}; \bm{\mu}_t^{(l)}, \textrm{diag} ((\bm{\sigma}_t^{(l)})^2))$, where $\bm{\mu}_t^{(l)}$ and $\bm{\sigma}_t^{(l)}$ are functions of $\mathbf{z}_{<t}^{(l)}$ and $\mathbf{z}_t^{(>l)}$. We note that \cite{kumar2019videoflow} parameterize $\bm{\mu}_t^{(l)}$ as $\bm{\mu}_t^{(l)} = \mathbf{z}_{t-1}^{(l)} + \widetilde{\bm{\mu}}_t^{(l)}$, where $\widetilde{\bm{\mu}}_t^{(l)}$ is the function. \cite{kumar2019videoflow} refer to this as \texttt{latent\_skip}. This is already a special case of an affine autoregressive flow, with a hard-coded shift of $\mathbf{z}_{t-1}^{(l)}$ and a scale of $\mathbf{1}$. We parameterize an affine autoregressive flow at each latent level, with a shift, $\bm{\alpha}_t^{(l)}$, and scale, $\bm{\beta}_t^{(l)}$, which are function of $\mathbf{z}_{<t}^{(l)}$, using the same 5-block ResNet architecture as \cite{kumar2019videoflow}. In practice, these functions are conditioned on the variables at the past three time steps. The affine autoregressive flow produces a new variable:
\begin{equation}
    \nonumber \mathbf{u}_t^{(l)} = \frac{\mathbf{z}_t^{(l)} - \bm{\alpha}_t^{(l)}}{\bm{\beta}_t^{(l)}},
\end{equation}
which we then model using the same prior distribution and architecture as \cite{kumar2019videoflow}: $p_\theta (\mathbf{u}_t^{(l)} | \mathbf{z}_{<t}^{(l)}, \mathbf{z}_t^{(>l)}) = \mathcal{N} (\mathbf{u}_t^{(l)}; \bm{\mu}_t^{(l)}, \textrm{diag} ((\bm{\sigma}_t^{(l)})^2))$, where $\bm{\mu}_t^{(l)}$ and $\bm{\sigma}_t^{(l)}$, again, are functions of $\mathbf{z}_{<t}^{(l)}$ (or, equivalently $\mathbf{u}_{<t}^{(l)}$) and $\mathbf{z}_t^{(>l)}$.

\subsection{Non-Video Sequence Modeling Architecture}
We again compare various model classes in terms of log-likelihood estimation. We use fully-connected networks to parameterize all functions within the prior, approximate posterior, and conditional likelihood of each model. All networks are $2$ layers of $256$ units with highway connectivity \citep{srivastava2015training}. For autoregressive flows, we use \texttt{ELU} non-linearities \citep{clevert2015fast}. For stability, we found it necessary to use \texttt{tanh} non-linearities in the networks for SLVMs (prior, conditional likelihood, and approximate posterior). In SLVMs, the prior is conditioned on $\mathbf{z}_{t-1}$, the approximate posterior is conditioned on $\mathbf{z}_{t-1}$ and $\mathbf{y}_{t}$, and the conditional likelihood is conditioned on $\mathbf{z}_t$. We use a latent space dimensionality of $16$ for all SLVMs.

\subsection{Training Set-Up}

We use the Adam optimizer \cite{kingma2014adam} with a learning rate of $1 \times 10^{-4}$ to train all the models. For Moving MNIST, we use a batch size of $16$ and train for $200,000$ iterations for SLVM and $100,000$ iterations for 1-AF, 2-AF and SLVM + 1-AF. For BAIR Robot Pushing, we use a batch size of $8$ and train for $200,000$ iterations for all models. For KTH Actions, we use a batch size of $8$ and train for $90,000$ iterations for all models. Batch norm \citep{ioffe2015batch} is applied to all convolutional layers that do not output distribution or affine transform parameters. We randomly crop sequences of length $13$ from all sequences and evaluate on the last $10$ frames. For AF-2 models, we crop sequences of length $16$ in order to condition both flows on three previous inputs. For VideoFlow experiments, we use the same hyper-parameters as \cite{kumar2019videoflow} (with the exception of the two architecture changes mentioned above) and train for $100,000$ iterations.

\begin{table}[h]
\begin{tabular}{lcccc}
\toprule
    Model & 1-AF & 2-AF & SLVM & SLVM + 1-AF \\
\midrule
    Moving Mnist & $343$k & $686$k & $11,302$k & $10,592$k  \\
\midrule
    BAIR Robot Pushing & $363$k & $726$k & $11,325$k & $10,643$k \\
\midrule
KTH Action & $343$k & $686$k & $11,302$k & $10,592$k  \\
    \bottomrule
\end{tabular}
\caption{\textbf{Number of parameters for each model on each dataset.} Flow-based models contain relatively few parameters as compared with the SLVM, as our flows consist primarily of $3 \times 3$ convolutions with limited channels. In the SLVM, we use two LSTM layers for modeling the prior and approx.~posterior distribution of the latent variable, while in SLVM + 1-AF, we use a single LSTM layer for each.}
\end{table}

\subsection{Quantifying Decorrelation}

To quantify the temporal redundancy reduction resulting from affine autoregressive pre-processing, we evaluate the empirical correlation between successive frames for the data observations and noise variables, averaged over spatial locations and channels. This is an average normalized version of the \textit{auto-covariance} of each signal with a time delay of $1$ time step. Specifically, we estimate the temporal correlation as
\begin{equation}
    \textrm{corr}_{\mathbf{x}} \equiv \frac{1}{HWC} \cdot \sum_{i, j, k}^{H, W, C} \mathbb{E}_{x^{(i, j, k)}_t , x^{(i, j, k)}_{t+1} \sim \mathcal{D}} \left[ \xi_{t, t+1} (i,j, k) \right],
    \label{eq: temp corr}
\end{equation}
where the term inside the expectation is
\begin{equation}
    \xi_{t, t+1} (i,j, k) \equiv \frac{(x^{(i, j, k)}_t - \mu^{(i, j, k)})(x^{(i, j, k)}_{t+1} - \mu^{(i, j, k)})}{\left(\sigma^{(i, j, k)}\right)^2}.
    \label{eq: temp corr}
\end{equation}
Here, $x_t^{(i, j, k)}$ denotes the image at location $(i, j)$ and channel $k$ at time $t$, $\mu^{(i, j, k)}$ is the mean of this dimension, and $\sigma^{(i, j, k)}$ is the standard deviation. $H, W,$ and $C$ respectively denote the height, width, and number of channels of the observations, and $\mathcal{D}$ denotes the dataset. We define an analogous expression for $\mathbf{y}$, denoted $\textrm{corr}_{\mathbf{y}}$.
\newpage
\section{Illustrative Example}
\label{sec: motivating example}

\begin{figure}[t!]
    \centering
    \includegraphics[width=\textwidth]{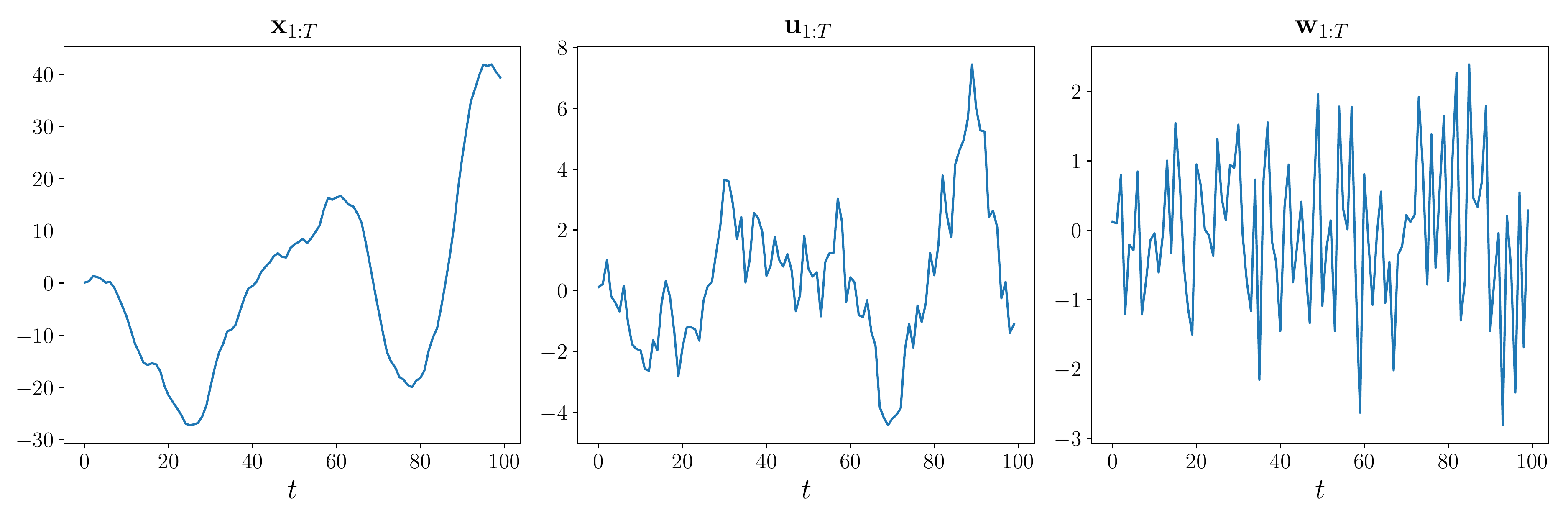}
    \caption{\textbf{Motivating Example}. Plots are shown for a sample of $\mathbf{x}_{1:T}$ (left),  $\mathbf{u}_{1:T}$ (center), and $\mathbf{w}_{1:T}$ (right).  Here, $\mathbf{w}_{1:T} \sim \mathcal{N} (\mathbf{w}_{1:T}; \mathbf{0}, \mathbf{I})$, and $\mathbf{u}$ and $\mathbf{x}$ are initialized at $0$. Moving from $\mathbf{x} \rightarrow \mathbf{u} \rightarrow \mathbf{w}$ via affine transforms results in successively less temporal correlation and therefore simpler dynamics.}
    \label{fig: simple example}
\end{figure}

To build intuition behind the benefits of temporal pre-processing (e.g., decorrelation) for downstream dynamics modeling, we present the following simple, kinematic example. Consider the discrete dynamical system defined by the following set of equations:
\begin{align}
    \mathbf{x}_t & = \mathbf{x}_{t-1} + \mathbf{u}_t, \label{eq: simple dynamics} \\
    \mathbf{u}_t & = \mathbf{u}_{t-1} + \mathbf{w}_t,
\end{align}
where $\mathbf{w}_t \sim \mathcal{N} (\mathbf{w}_t; \mathbf{0}, \bm{\Sigma})$. We can express $\mathbf{x}_t$ and $\mathbf{u}_t$ in probabilistic terms as
\begin{align}
    \mathbf{x}_t & \sim \mathcal{N} (\mathbf{x}_t; \mathbf{x}_{t-1} + \mathbf{u}_{t-1}, \bm{\Sigma}), \\
    \mathbf{u}_t & \sim \mathcal{N} (\mathbf{u}_t; \mathbf{u}_{t-1}, \bm{\Sigma}). \label{eq: simple u dynamics}
\end{align}
Physically, this describes the noisy dynamics of a particle with momentum and mass $1$, subject to  Gaussian noise. That is, $\mathbf{x}$ represents position, $\mathbf{u}$ represents velocity, and $\mathbf{w}$ represents stochastic forces. If we consider the dynamics at the level of $\mathbf{x}$, we can use the fact that $\mathbf{u}_{t-1} = \mathbf{x}_{t-1} - \mathbf{x}_{t-2}$ to write
\begin{equation}
    p (\mathbf{x}_t | \mathbf{x}_{t-1}, \mathbf{x}_{t-2}) = \mathcal{N} (\mathbf{x}_t; \mathbf{x}_{t-1} + \mathbf{x}_{t-1} - \mathbf{x}_{t-2}, \bm{\Sigma}).
\end{equation}
Thus, we see that in the space of $\mathbf{x}$, the dynamics are second-order Markov, requiring knowledge of the past two time steps. However, at the level of $\mathbf{u}$ (Eq.\ \ref{eq: simple u dynamics}), the dynamics are first-order Markov, requiring only the previous time step. Yet, note that $\mathbf{u}_t$ is, in fact, an affine autoregressive transform of $\mathbf{x}_t$ because $\mathbf{u}_t = \mathbf{x}_t - \mathbf{x}_{t-1}$ is a special case of the general form $\frac{\mathbf{x}_t - \bm{\mu}_\theta (\mathbf{x}_{<t})}{\bm{\sigma}_\theta (\mathbf{x}_{<t})}$. In Eq.\ \ref{eq: simple dynamics}, we see that the Jacobian of this transform is $\partial \mathbf{x}_t / \partial \mathbf{u}_t = \mathbf{I}$, so, from the change of variables formula, we have $p (\mathbf{x}_t | \mathbf{x}_{t-1}, \mathbf{x}_{t-2}) = p (\mathbf{u}_t | \mathbf{u}_{t-1})$. In other words, an affine autoregressive transform has allowed us to convert a second-order Markov system into a first-order Markov system, thereby simplifying the dynamics. Continuing this process to move to $\mathbf{w}_t = \mathbf{u}_t - \mathbf{u}_{t-1}$, we arrive at a representation that is entirely temporally decorrelated, i.e. no dynamics, because $p (\mathbf{w}_t) = \mathcal{N} (\mathbf{w}_t; \mathbf{0}, \bm{\Sigma})$. A sample from this system is shown in Figure \ref{fig: simple example}, illustrating this process of temporal decorrelation.
\newpage

\section{Additional Experimental Results}
\label{appendix: add exp res}

\begin{table}[h]
\begin{tabular}{@{\hskip2pt}p{2.7cm}@{\hskip18pt}rrrr}
\toprule
 & M-MNIST & BAIR & KTH  \\
\midrule
1-AF      & $2.06$  &  $2.98$ & $2.95$ \\
2-AF      & $2.04$  &  $2.76$ & $2.95$ \\
\midrule
SLVM      & $\leq 1.93$  &  $\leq 3.46$ & $\leq 3.05$ \\
SLVM + $\Delta \mathbf{x}$   & $\leq 2.47$  & $\leq 3.05$  & $\leq 2.46$ \\
SLVM + 1-AF      & $\leq \mathbf{1.85}$  &  $\leq \mathbf{2.31}$ & $\leq \mathbf{2.21}$ \\
\midrule
VF & -- & $1.50$ & -- \\
VF + AF & -- & $\mathbf{1.49}$ & -- \\
\bottomrule
\end{tabular}
\caption{{\bf Training Quantitative Comparison.} Average training negative log-likelihood in \textit{nats per dim.} for Moving MNIST, BAIR Robot Pushing, and KTH Actions.}
\label{table:elbo_comparison_train}
\end{table}

\subsection{Additional Qualitative Results}
\label{subsec: add qual res}

\begin{figure}[H]
    \centering
    \begin{subfigure}[t]{0.49\textwidth}
        \centering
        \includegraphics[width=\textwidth]{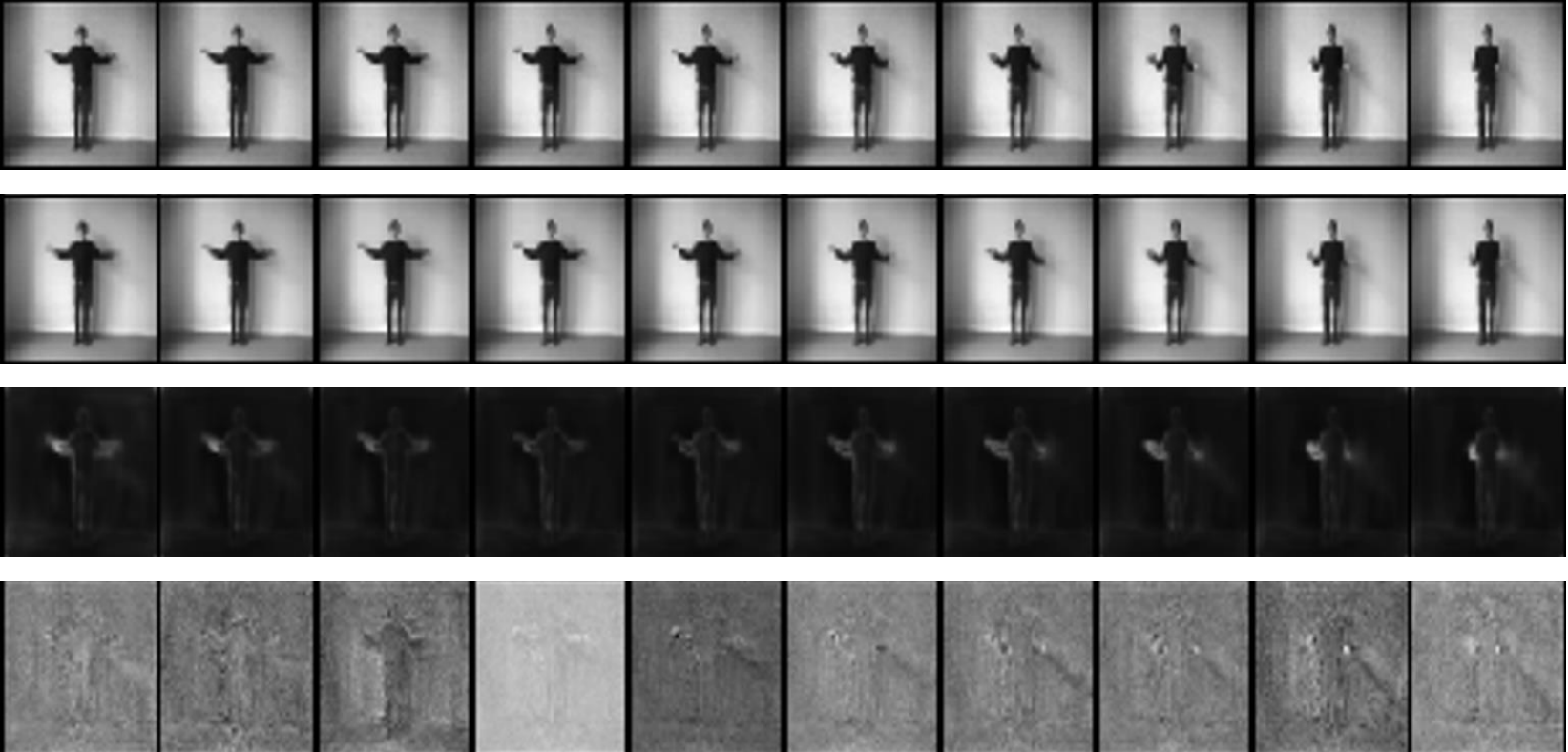}
        \caption{}
        \label{fig: flow vis kth}
    \end{subfigure}%
    ~ 
    \begin{subfigure}[t]{0.49\textwidth}
        \centering
        \includegraphics[width=\textwidth]{kth_latent_flow_visualization.pdf}
        \caption{}
        \label{fig: latent + flow vis kth}
    \end{subfigure}
    \caption{\textbf{Autoregressive Flow Visualization on KTH Action}. Visualization of the flow component for \textbf{(a)} standalone flow-based models and \textbf{(b)} sequential latent variable models with flow-based conditional likelihoods for KTH Actions. From top to bottom, each figure shows 1) the original frames, $\mathbf{x}_t$, 2) the predicted shift, $\bm{\mu}_\theta (\mathbf{x}_{<t})$, for the frame, 3) the predicted scale, $\bm{\sigma}_\theta (\mathbf{x}_{<t})$, for the frame, and 4) the noise, $\mathbf{y}_t$, obtained from the inverse transform.}
    \label{fig: qualitative visualization 2}
\end{figure}

\begin{figure}[H]
    \centering
    \includegraphics[width=\textwidth]{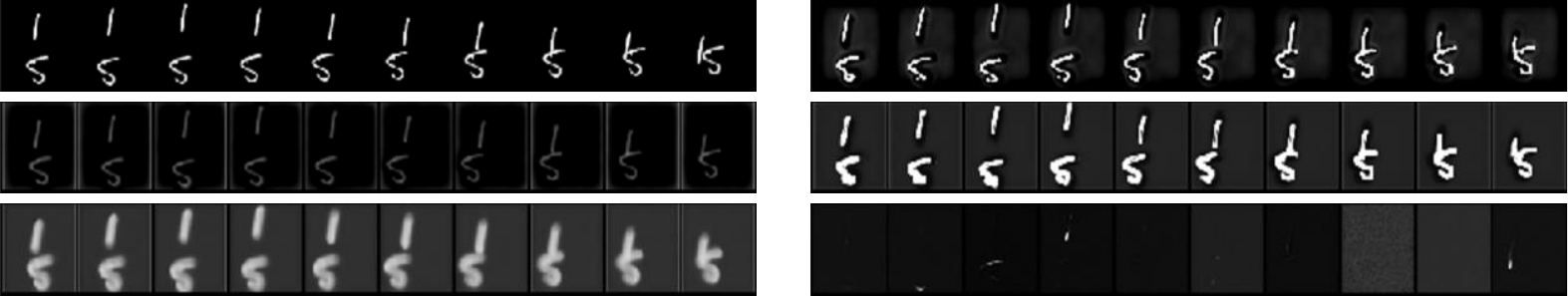}
    \caption{\textbf{SLVM w/ 2-AF Visualization on Moving MNIST}. Visualization of the flow component for sequential latent variable models with 2-layer flow-based conditional likelihoods for Moving MNIST. From top to bottom on the left side, each figure shows 1) the original frames, $\mathbf{x}_t$, 2) the lower-level predicted shift, $\bm{\mu}_\theta^1 (\mathbf{x}_{<t})$, for the frame, 3) the predicted scale, $\bm{\sigma}_\theta^1 (\mathbf{x}_{<t})$, for the frame. On the right side, from top to bottom, we have 1) the higer-level predicted shift, $\bm{\mu}_\theta^2 (\mathbf{x}_{<t})$, for the frame, 3) the predicted scale, $\bm{\sigma}_\theta^2 (\mathbf{x}_{<t})$, for the frame and 4) the noise, $\mathbf{y}_t$, obtained from the inverse transform.}
    \label{fig:mnist_2flow_rmn}
\end{figure}

\begin{figure}[t]
    \centering
    \begin{subfigure}[h]{0.49\textwidth}
        \centering
        \includegraphics[width=\textwidth]{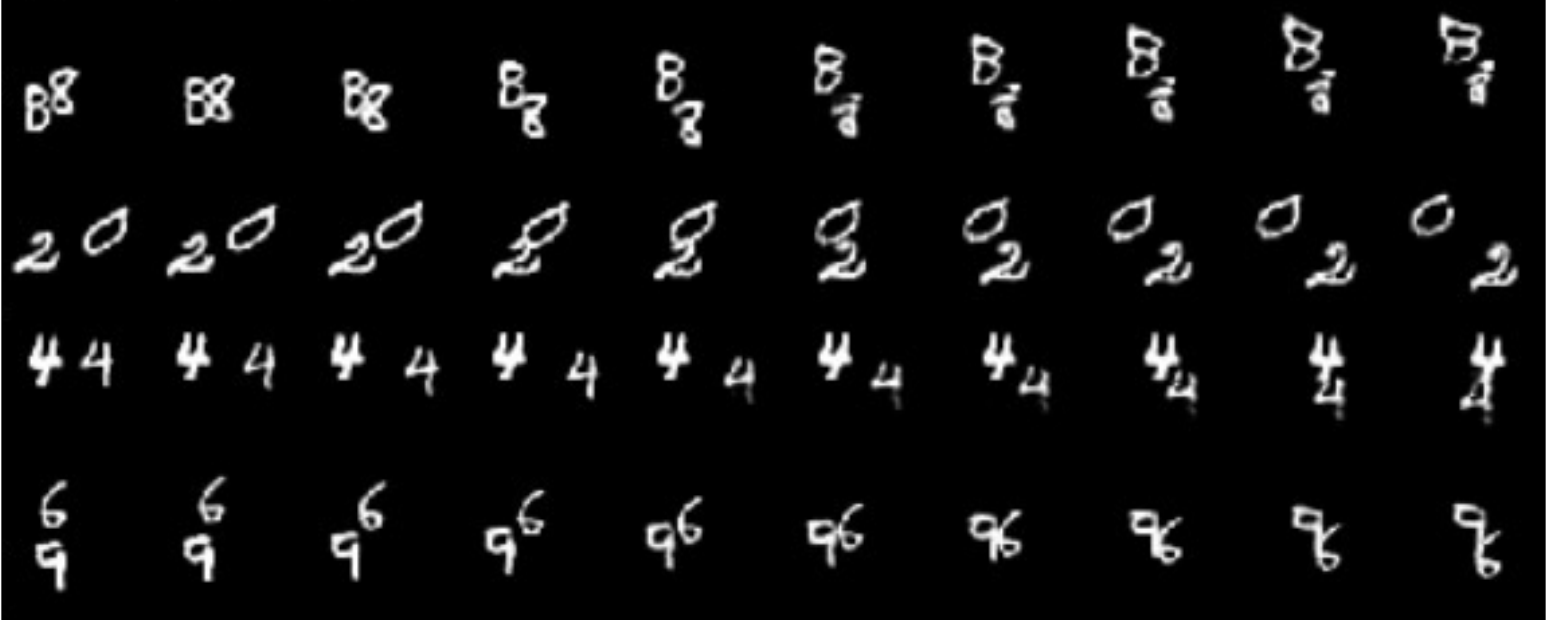}
        \label{fig: mnist gen1}
    \end{subfigure}%
    ~ 
    \begin{subfigure}[h]{0.49\textwidth}
        \centering
        \includegraphics[width=\textwidth]{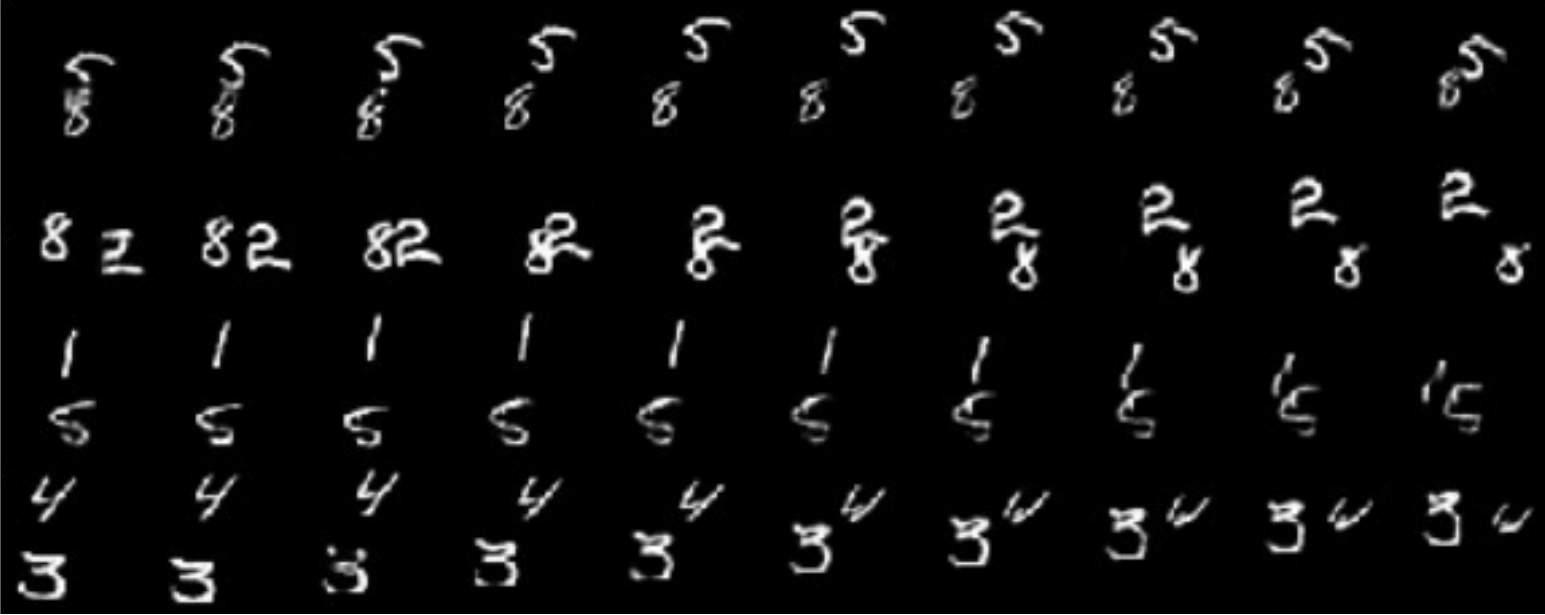}
        \label{fig: mnist gen2}
    \end{subfigure}
    \caption{\textbf{Generated Moving MNIST Samples}. Sample frame sequences generated from a 2-AF model.}
    \label{fig: mnist gen}
\end{figure}

\begin{figure}[H]
    \centering
    \includegraphics[width=0.9\textwidth]{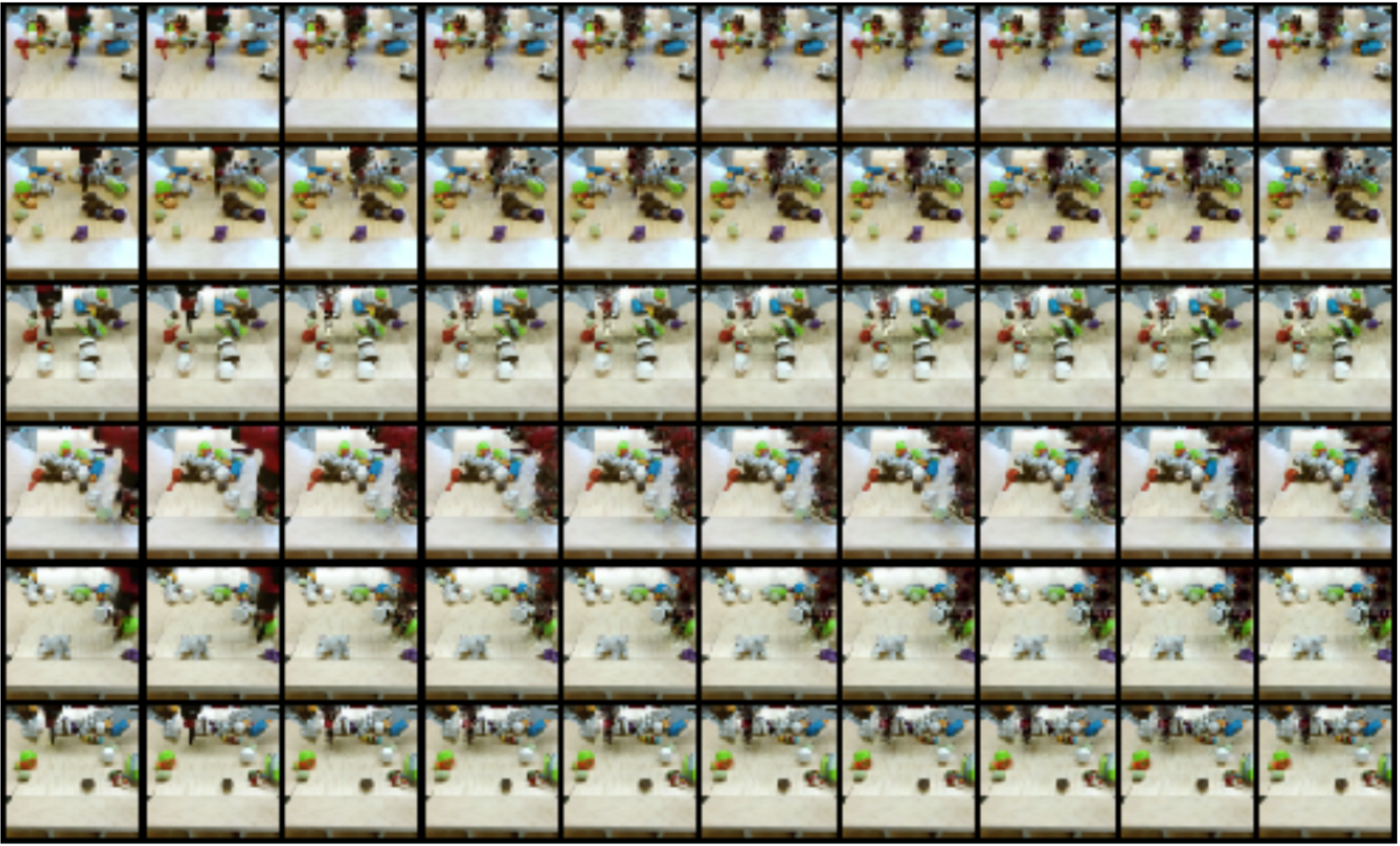}
    \caption{\textbf{Generated BAIR Robot Pushing Samples}. Sample frame sequences generated from SLVM + 1-AF. Sequences remain relatively coherent throughout, but do not display large changes across frames.}
    \label{fig:bair gen}
\end{figure}

    
\end{document}